\documentclass{article}

\PassOptionsToPackage{numbers, compress}{natbib}

\usepackage[nonatbib,preprint]{neurips_2023}
\usepackage{authblk}
\usepackage[normalem]{ulem}

\usepackage{setspace}
\usepackage{comment}
\usepackage{amsmath,amsthm,amssymb}
\usepackage{ulem}
\usepackage{makecell}

\newcommand{\bs}{\boldsymbol}

\providecommand{\keywords}[1]
{\textbf{\text{Keywords: }} #1}
\newcommand{\ee}{\end{equation}}
\newcommand{\be}{\begin{equation}}
\newcommand{\ec}{\end{center}}
\newcommand{\bc}{\begin{center}}
\newcommand{\eea}{\end{eqnarray}}
\newcommand{\bea}{\begin{eqnarray}}
\newcommand{\bd}{\begin{description}}
\newcommand{\ed}{\end{description}}
\newcommand{\bi}{\begin{itemize}}
\newcommand{\ei}{\end{itemize}}

\newcommand{\bt}{\bs{\theta}}

\newcommand{\bmx}{\bm{x}}

\usepackage{commath}
\usepackage{mathtools}
\usepackage{algorithm,algpseudocode}
\usepackage{hyperref}
\usepackage{color}
\usepackage{amsfonts}

\usepackage{diagbox}
\usepackage{bm}
\usepackage{booktabs}       
\usepackage{multirow}
\usepackage{subfigure}
\usepackage[percent]{overpic}
\usepackage{caption}

\algnewcommand{\Inputs}[1]{%
  \State \textbf{Inputs:}
  \Statex \hspace*{\algorithmicindent}\parbox[t]{.8\linewidth}{\raggedright #1}
}
\algnewcommand{\Initialize}[1]{%
  \State \textbf{Initialize:}
  \Statex \hspace*{\algorithmicindent}\parbox[t]{.8\linewidth}{\raggedright #1}
}
\algnewcommand{\Outputs}[1]{%
  \State \textbf{Outputs:}
  \Statex \hspace*{\algorithmicindent}\parbox[t]{.8\linewidth}{\raggedright #1}
}

\title{A unified physics-informed generative operator framework for general inverse problems}

\author[a]{Gang Bao}
\author[b*]{Yaohua Zang}
\affil[a]{Zhejiang University, Center for Interdisciplinary Applied Mathematics, Yuhangtang Road 886, 310058 Hangzhou, China}
\affil[b*]{Technical University of Munich, Professorship of Data-driven Materials Modeling, School of Engineering and Design, Boltzmannstr. 15, 85748 Garching, Germany}
\affil[*]{\text{yaohua.zang@tum.de}}
\begin{document}
\maketitle
\begin{abstract}
Solving inverse problems governed by partial differential equations (PDEs) is central to science and engineering, yet remains challenging when measurements are sparse, noisy, or when the underlying coefficients are high-dimensional or discontinuous. Existing deep learning approaches either require extensive labeled datasets or are limited to specific measurement types, often leading to failure in such regimes and restricting their practical applicability.
Here, a novel generative neural operator framework, IGNO, is introduced to overcome these limitations. IGNO unifies the solution of inverse problems from both point measurements and operator-valued data without labeled training pairs. This framework encodes high-dimensional, potentially discontinuous coefficient fields into a low-dimensional latent space, which drives neural operator decoders to reconstruct both coefficients and PDE solutions. Training relies purely on physics constraints through PDE residuals, while inversion proceeds via efficient gradient-based optimization in latent space, accelerated by an a priori normalizing flow model.
Across a diverse set of challenging inverse problems, including recovery of discontinuous coefficients from solution-based measurements and the EIT problem with operator-based measurements, IGNO consistently achieves accurate, stable, and scalable inversion even under severe noise. It consistently outperforms the state-of-the-art method under varying noise levels and demonstrates strong generalization to out-of-distribution targets. These results establish IGNO as a unified and powerful framework for tackling challenging inverse problems across computational science domains.
\end{abstract}
\section{Introduction}\label{sec:introduction}
Inverse problems arise whenever unknown properties of a system need to be inferred from indirect or incomplete observations of its physical states. They are fundamental to many areas of science and engineering, from reconstructing subsurface properties in seismic imaging \cite{stolt2012seismic} and determining tissue conductivity in medical diagnostics \cite{uhlmann2009electrical,bertero2021introduction} to discovering microstructural properties in materials design \cite{zang2025psp}. These problems are governed by partial differential equations (PDEs) encoding physical laws relating unknown parameters to observable quantities \cite{hahn2012heat,white2006viscous,van2007electromagnetic,ogden1997non}. However, PDE inverse problems are notoriously challenging: they are typically high-dimensional, ill-posed, and sensitive to noise. Even small measurement errors or model uncertainties can lead to large deviations in reconstructed parameters, making robust and efficient inversion a long-standing challenge.

Because analytical solutions are rarely attainable except in oversimplified cases, traditional approaches formulate inverse problems as constrained optimization tasks that iteratively update parameters while satisfying PDE constraints through forward and adjoint solves \cite{chavent2010nonlinear,byrne2014iterative}. While mathematically rigorous, these methods demand hundreds of expensive PDE simulations per inversion, making them impractical for large-scale or high-dimensional problems. Surrogate and reduced-order models \cite{frangos2010surrogate} reduce cost but typically sacrifice accuracy beyond their calibration regimes. Bayesian inference frameworks \cite{calvetti2018inverse} provide uncertainty quantification but face the curse of dimensionality, as sampling high-dimensional posteriors is computationally prohibitive.

Physics-informed machine learning (PIML) has recently introduced a paradigm shift by embedding governing physical laws directly into neural network training \cite{han2018solving,sirignano2018dgm,yu2018deep,raissi2019physics,kharazmi2019variational,zang2020weak,zang2023particlewnn}. For inverse problems, PIML treats the unknown coefficients as trainable parameters while enforcing physical laws, enabling simultaneous recovery of hidden parameters and solutions \cite{raissi2019physics,bao2020numerical,chen2020physics,rasht2022physics,gao2022physics,jagtap2022physics,scholz2025weak}. However, existing PIML-based inversion methods exhibit notable limitations. They are primarily restricted to \textbf{solution-based inverse problems} with pointwise measurements and struggle fundamentally with \textbf{operator-based inverse problems}, such as electrical impedance tomography (EIT) \cite{uhlmann2009electrical}, where measurements consist of input–output operators like Dirichlet-to-Neumann (DtN) maps rather than scalar values. Inverting such operators requires solving entire families of PDEs under multiple boundary conditions, dramatically increasing computational costs. Moreover, PIML training remains expensive, and the lack of structured representations for high-dimensional parameter spaces leads to highly non-convex optimization landscapes and poor convergence.

Deep neural operators (DNOs) \cite{li2020neural,li2020fourier,lu2021learning,kovachki2023neural} have emerged as a promising alternative, capable of learning mappings between infinite-dimensional function spaces and predicting PDE solutions across entire families of inputs after a single training phase. DNO-based inversion typically follows two strategies: (i) using trained DNOs as fast forward surrogates within iterative optimization \cite{gao2024adaptive,guo2024reduced}, which still requires costly searches in high-dimensional coefficient spaces; or (ii) training inverse DNOs that directly map observations to coefficients \cite{molinaro2023neural,wang2024latent,long2024invertible}, which demand massive labeled datasets, are limited to fixed sensor configurations, and exhibit poor robustness to noise and distribution shifts. 

Recent advances in physics-informed DNOs (PI-DNOs) have sought to overcome these challenges by embedding PDE constraints into training, thereby reducing data dependence \cite{wang2021learning,li2024physics,navaneeth2024physics,jiao2024solving,zang2025dgenno}. For example, PI-DeepONet \cite{jiao2024solving} applies Bayesian MCMC for inversion but struggles in high-dimensional regimes and cannot handle operator-based measurements. PI-DIONs \cite{cho2024physics} trains via strong-form residuals but is restricted to fixed sensors and requires extensive solution-based measurements. DGenNO \cite{zang2025dgenno} introduces latent input compression for Darcy flow but remains confined to solution-based settings and exhibits sensitivity to initialization.

\begin{figure}[!htbp]
\centering
\includegraphics[width=0.9\textwidth]{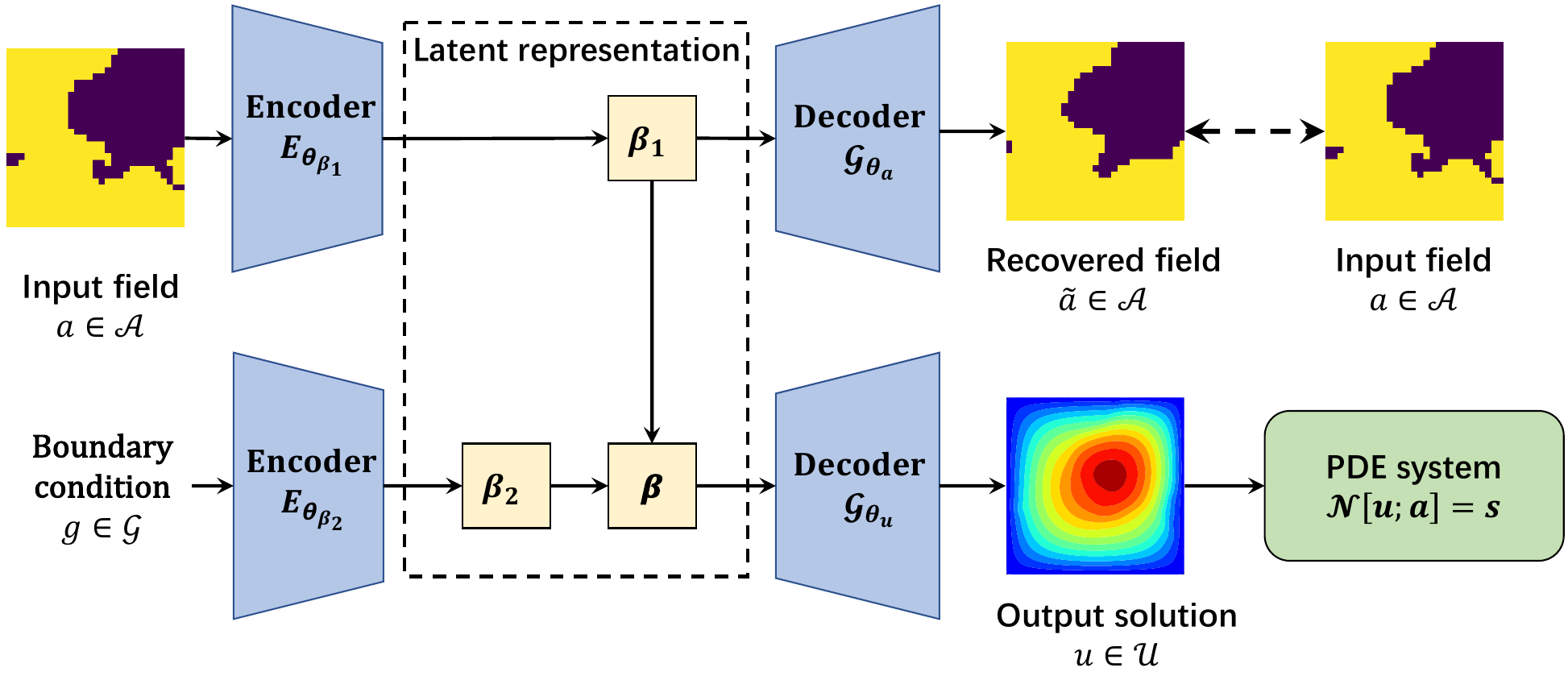}
\caption{The IGNO framework for PDE inverse problems. The generative neural operator architecture encodes the coefficient field $a$ and the boundary condition $g$ into a structured latent variable $\bm{\beta} = (\bm{\beta}_1, \bm{\beta}_2)$ via encoders $E_{\bt_{\bm{\beta}_1}}$ and $E_{\bt_{\bm{\beta}_2}}$, respectively. Two MultiONet-based decoders reconstruct the coefficient $a$ from $\bm{\beta}_1$ (coefficient decoder $\mathcal{G}_{\bt_a}$) and predict the PDE solution $u$ from $\bm{\beta}$ (solution decoder $\mathcal{G}_{\bt_u}$). The model is trained in a purely physics-informed manner, driven solely by governing PDE residuals without requiring paired input-output data. For inversion, gradient-based optimization in the low-dimensional latent space enables efficient recovery of unknown coefficients from sparse, noisy observations.}
\label{fig:model}
\end{figure}

To address these limitations, this study proposes the \textbf{Inverse Generative Neural Operator (IGNO)}, a unified, physics-informed generative framework designed to efficiently and robustly solve general PDE inverse problems (Fig. \ref{fig:model}). The key innovation of IGNO lies in transforming the intractable task of optimizing directly within high-dimensional, irregular coefficient spaces into efficient gradient-based optimization within a smooth, low-dimensional latent manifold. This is achieved by the generative architecture of IGNO, which consists of dual encoders that compress coefficient fields and boundary conditions into compact latent representations, $\bm{\beta}=(\bm{\beta}_1,\bm{\beta}_2)$, and two MultiONet-based decoders \cite{zang2025dgenno} that respectively reconstruct coefficients from $\bm{\beta}_1$ and predict PDE solutions from $\bm{\beta}$. Training of IGNO proceeds in a purely physics-informed manner by minimizing PDE residuals formulated in weak or strong form, requiring no labeled input–output pairs or precomputed forward simulations. This self-supervised paradigm enables efficient learning even with minimal data, while ensuring consistency with governing physical laws. During inversion, IGNO performs gradient-based optimization directly in the latent space to minimize discrepancies between predictions and observations while maintaining PDE consistency. To enhance convergence and robustness, a normalizing flow is also integrated into IGNO that maps the latent space to a standard Gaussian distribution, providing statistically informed initializations that accelerate convergence and stabilize optimization in noisy or ill-posed settings. A critical distinguishing capability of IGNO is its unified treatment of diverse measurement modalities within a single framework. Through its dual-encoder architecture, IGNO naturally accommodates both solution-based observations consisting of sparse point measurements and operator-based measurements such as DtN maps in EIT. For operator-valued data, IGNO handles multiple boundary conditions simultaneously without repeatedly solving PDEs during training, enabling efficient inversion from operator-based measurements where existing PI-DNOs struggle to operate.

The framework is validated through extensive numerical experiments encompassing both continuous and discontinuous inverse targets. For solution-based problems, IGNO reconstructs smooth and piecewise-constant permeability fields in Darcy flow from sparse pressure measurements. For operator-based problems, it solves the EIT problem by recovering conductivity from DtN boundary operators. Across varying noise levels, out-of-distribution targets, and high-dimensional coefficient spaces, IGNO achieves reconstruction errors 3 to 6 times lower than the state-of-the-art method, with superior recovery of fine-scale features and strong extrapolation performance. Notably, IGNO maintains stable and accurate inversion in challenging scenarios where the baseline method produces substantial distortions or fails entirely, including recovery of discontinuous coefficients where traditional gradient-based methods cannot operate due to undefined gradients.
These results establish IGNO as a scalable and broadly applicable framework for PDE inverse problems. By synergistically combining generative latent modeling, neural operator architectures, and physics-informed learning, IGNO effectively addresses key challenges in data efficiency, noise robustness, and generalization that have long hindered inverse problem solvers. Beyond the canonical problems demonstrated here, the methodology can naturally extend to inverse problems in geophysical exploration, materials characterization, and climate modeling, illustrating how the principled fusion of generative modeling, operator learning, and physics constraints can advance the frontier of computational science.

\section{Results}\label{sec:results}
\subsection{Validation strategy and experimental setup}
We validate IGNO across a suite of canonical PDE inverse problems that collectively test the framework’s ability to handle diverse coefficient structures, measurement modalities, and noise levels. The tested problems include the recovery of (i) continuous coefficients and (ii) piecewise-constant coefficients in Darcy flow, and (iii) the reconstruction of electrical conductivity in the EIT problem. These examples progressively increase in difficulty from low-noise, in-distribution inference to highly noisy, out-of-distribution recovery from operator-valued data, allowing systematic evaluation of IGNO's accuracy, robustness, and generalization.
In all experiments, IGNO is compared with the state-of-the-art Physics-Informed Deep Inverse Operator Networks (PI-DIONs) \cite{cho2024physics} under consistent training and optimization setups, except in the EIT case. Details of the PI-DIONs implementation are provided in Note 4 of the Supplementary Information (SI). We note that PI-DIONs cannot handle operator-based measurements, thus limiting comparisons to solution-based cases. For each problem, we evaluate performance under three noise levels: low (SNR = 50 dB), medium (SNR = 25 dB), and high (SNR = 15 dB), where SNR denotes the signal-to-noise ratio. Quantitative evaluations use the \textbf{root-mean-square error (RMSE)} for continuous targets and the \textbf{cross-correlation indicator ($I_{corr}$)} for piecewise-constant targets, complemented by visual assessment of fine-scale reconstructions.

Training data consist of 1,000 coefficient samples for the Darcy flow problem with continuous permeabilities and 10,000 samples with piecewise-constant permeabilities. For the EIT problem, 1,000 conductivity fields and 20 boundary conditions are used, leading to a total of $1000\times20$ samples. Crucially, training of IGNO requires only coefficient fields and boundary conditions, without any precomputed PDE solutions, whereas the PI-DIONs method relies on finite element method (FEM) solutions for all training samples. Finally,  the normalizing flow model used for latent-space initialization is trained on the learned latent representations after the main model training. Architectural and optimization details are summarized in the \textbf{Methods} or \textbf{Supplementary Methods} in SI.

\subsection{Continuous coefficient recovery from solution-based measurements}
\begin{figure}[!htbp]
\centering
\includegraphics[width=0.6\textwidth]{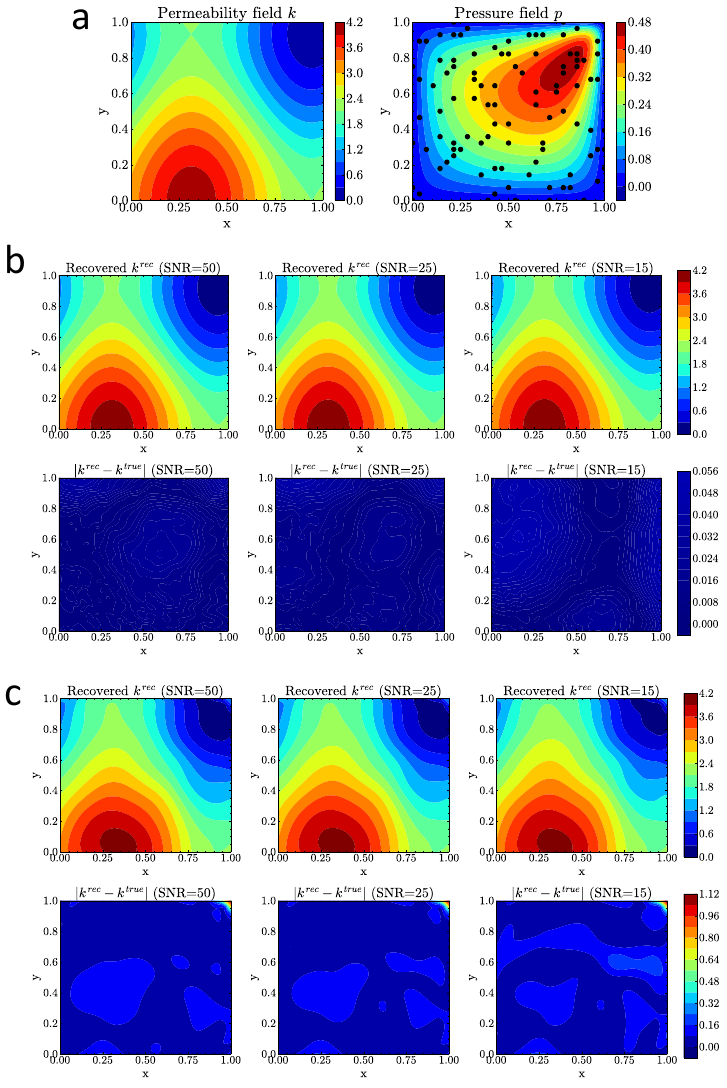}
\caption{Recovery of \textbf{continuous coefficients} with solution-based measurements (\textbf{In-distribution case}): (a) the ground truth permeability field $k$ and the corresponding pressure field $p$ (black dots denote $m=100$ random sensors); (b) recovered permeability $k^{rec}$ and corresponding pointwise absolute errors obtained by \textbf{IGNO} under different noise levels; (c) recovered permeability $k^{rec}$ and corresponding pointwise absolute errors obtained by \textbf{PI-DIONs} under different noise levels.}
\label{fig:smh_in}
\end{figure}

The first benchmark involves recovering a smooth permeability field in the Darcy flow equation from sparse, noisy pressure measurements. This problem serves as a controlled setting for evaluating IGNO’s data efficiency and noise robustness. The governing PDE is a second-order elliptic equation on the unit square domain:
\begin{equation}\label{eq:darcy_smh}
\begin{aligned}
-\nabla(k(x,y)\nabla p(x,y)) &= f(x,y), \quad \text{in}\ \Omega=[0,1]^2, \\
p(x,y) &= 0,\quad \text{on}\ \partial\Omega,
\end{aligned}
\end{equation}
where $k$ denotes the unknown permeability, $p$ the pressure, and $f = 10$ the source term. The permeability fields follow
$k(x,y) = 2.1 + \sin(\omega_1 x) + \cos(\omega_2 y)$, with $(\omega_1, \omega_2)$ sampled independently from the uniform distribution $\text{U}(0,7\pi/4)^2$. Measurements consist of pressure values at $m=100$ randomly placed interior sensors, shown as black dots in the right of Fig.~\ref{fig:smh_in}(a), and are corrupted by additive Gaussian noise at the specified SNR levels. 

Fig.\ref{fig:smh_in}(b)–(c) presents the reconstructed permeability fields and corresponding pointwise absolute errors for an in-distribution test case under three noise levels. Comparative performance results of both methods across different noise levels are presented in Supplementary Table 1 of the SI. As shown in Fig.\ref{fig:smh_in}(b), IGNO accurately recovers both the global patterns and fine-scale variations of the permeability field, even under severe noise (SNR = 15 dB), yielding smooth reconstructions with errors primarily localized in regions of rapid spatial variation. Reconstruction quality degrades only slightly as noise increases, with RMSEs of 0.56\%, 0.61\%, and 1.15\% for SNR values of 50, 25, and 15 dB, respectively. In contrast, PI-DIONs exhibit substantially higher errors across all noise levels, with RMSEs of 3.66\%, 3.80\%, and 4.15\% for SNR values of 50, 25, and 15 dB, respectively. As shown in Fig.~\ref{fig:smh_in}(c), the PI-DIONs method fails to resolve fine-scale structures and produces noticeable distortions, particularly in regions of low permeability. Pointwise absolute error maps further confirm that IGNO maintains uniformly small errors across the domain, whereas PI-DIONs exhibit larger, spatially correlated error patterns.

The superior performance of IGNO can be attributed to its latent-space optimization strategy. Instead of directly optimizing over the high-dimensional coefficient space, IGNO performs inversion within a compact, well-structured latent manifold where gradients are smooth and correlated with meaningful physical variations. This transformation substantially alleviates the ill-posedness inherent to direct inversion. The embedded PDE residual constraints further enhance stability by enforcing physical consistency between the reconstructed permeability and the corresponding solution. In addition, the a priori normalizing flow provides statistically informed initialization, which reduces the likelihood of convergence to poor local minima and improves overall optimization reliability. In contrast, the PI-DIONs method predicts coefficients directly from noisy observations in the original, unstructured parameter space. This direct mapping is highly sensitive to measurement noise and lacks an explicit mechanism to enforce physical consistency, often leading to reconstructions that deviate significantly from true fields.

\subsection{Discontinuous coefficient recovery from solution-based measurements}
\begin{figure}[!htbp]
\centering
\includegraphics[width=0.6\textwidth]{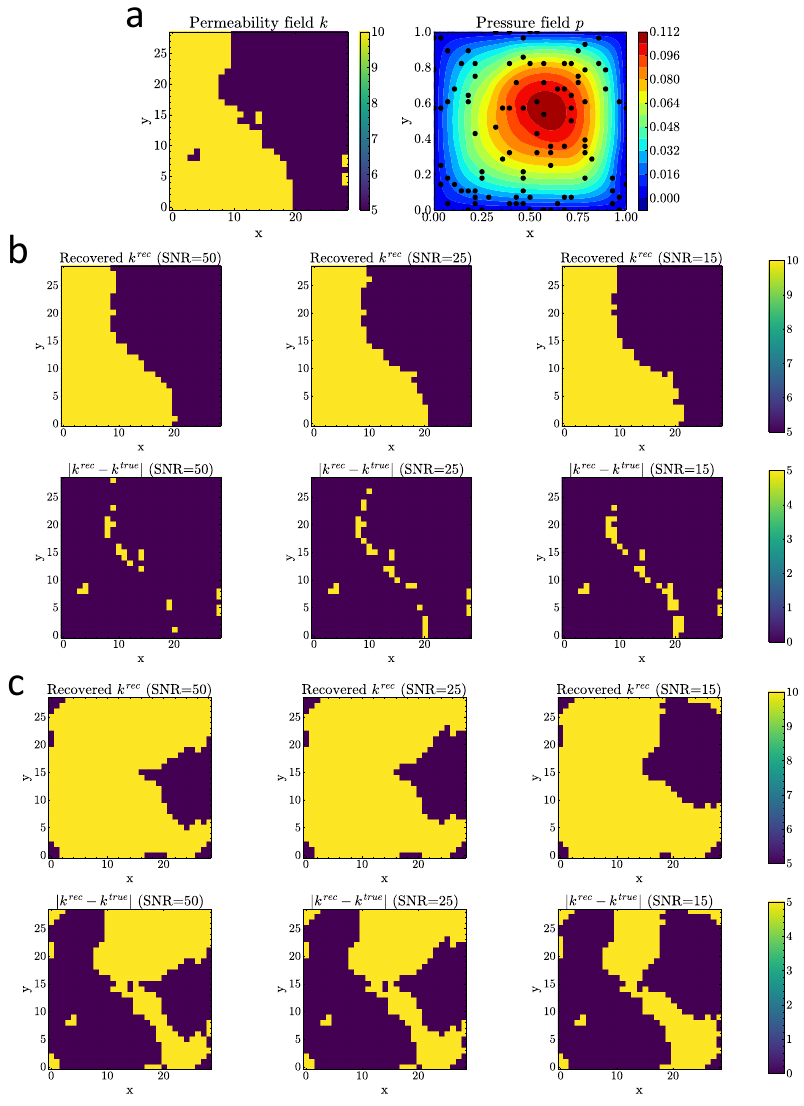}
\caption{Recovery of \textbf{piecewise-constant coefficients} with solution-based measurements (\textbf{In-distribution case}): (a) the true permeability field $k$ (Phase 1 shown in yellow) and the corresponding pressure field $p$ (black dots denote $m=100$ random sensors); (b) recovered permeability $k^{rec}$ and corresponding pointwise absolute errors obtained by IGNO under different noise levels; (c) recovered permeability $k^{rec}$ and corresponding pointwise absolute errors obtained by PI-DIONs under different noise levels.}
\label{fig:pwc_in}
\end{figure}

The second benchmark tackles a more challenging scenario in which the coefficient field is piecewise-constant, representing heterogeneous media with sharp interfaces between two permeability phases ($k = 10$ in phase 1 and $k = 5$ in phase 2), as shown in Fig.~\ref{fig:pwc_in}(a). This configuration closely mimics realistic geological or material systems where discontinuities naturally occur. This problem is particularly challenging because the gradients of $k$ are undefined at discontinuities, rendering traditional gradient-based optimization infeasible without special treatment. To overcome this challenge, IGNO employs a probabilistic phase decoder that estimates, for each spatial location, the likelihood of belonging to a given phase (phase 1). The decoder is trained using a cross-entropy loss instead of the mean-squared error (MSE) loss, as detailed in the SI. This probabilistic formulation prevents optimization breakdown at discontinuities and promotes naturally regularized reconstructions. During inference, a threshold of 0.5 is applied to the predicted probability map to obtain the final binary phase segmentation. For training, piecewise-constant permeability fields are generated from a cutoff Gaussian process $\mathcal{GP}(0,(-\Delta + 9I)^{-2})$ \cite{li2020fourier,zang2025dgenno}, and only $m = 100$ sparse pressure measurements are used for inversion, shown as black dots in the right of Fig.~\ref{fig:pwc_in}(a).

Fig.~\ref{fig:pwc_in}(b)–(c) presents the reconstructed permeability fields and pointwise absolute error maps obtained by both methods under varying noise levels for an in-distribution test case. The performance of both methods under different noise levels is reported in Supplementary Table 3 of the SI. Again, IGNO accurately reconstructs both the global phase topology and the sharp interfaces between phases across all noise levels, achieving cross-correlation indicators $I_{corr}$ of 0.969, 0.960, and 0.951 for SNR values of 50, 25, and 15 dB, respectively. As shown in Fig.~\ref{fig:pwc_in}(b), IGNO maintains morphologically faithful reconstructions even when measurements are contaminated with severe 15 dB noise, with pointwise absolute errors primarily confined to phase boundaries where discontinuities pose inherent challenges. In contrast, the PI-DIONs method fails to recover the correct structures, yielding $I_{corr}$ values of only 0.718, 0.715, and 0.726, indicative of largely random reconstructions with no resemblance to the ground truth. As seen in Fig.~\ref{fig:pwc_in}(c), the PI-DIONs method produces spatially incoherent patterns with widespread errors across the domain rather than localized near interfaces. This failure stems from two intrinsic limitations of PI-DIONs: (1) the reliance on strong-form PDE residuals, which require differentiability of $k$ and thus break down at discontinuities; and (2) direct recovery of $k$ in a high-dimensional, irregular function space, which makes the inversion process extremely challenging.
IGNO overcomes these challenges through its generative latent-space formulation. By embedding discontinuous coefficient fields into a smooth latent manifold, IGNO converts a fundamentally discrete inversion task into a differentiable, well-conditioned optimization problem. The trained coefficient decoder then maps the optimized latent representation back to the discontinuous field, effectively reconstructing sharp interfaces without requiring explicit gradient information. Furthermore, the normalizing flow–based initialization accelerates convergence and enhances reconstruction fidelity by providing latent variables corresponding to statistically plausible phase configurations.

\subsection{Recovery of conductivity from operator-based measurements: the EIT problem}
The EIT problem represents a fundamentally different class of inverse problems, where measurements correspond not to pointwise solution values but to an operator that maps boundary inputs to boundary responses. Specifically, the goal is to recover the spatially varying conductivity field $\gamma>0$ in the elliptic PDE:
\begin{equation}\label{eq:EIT}
\begin{aligned}
-\nabla\cdot(\gamma(x,y)\nabla u(x,y)) &= 0, \quad \text{in} \ \Omega=[0,1]^2, \\
u(x,y) &= g(x,y),\quad \text{on}\ \partial\Omega,
\end{aligned}
\end{equation}
from the DtN operator $\Lambda_\gamma[g]: g \mapsto \gamma\frac{\partial u}{\partial\vec{n}}\big|_{\partial\Omega}$, which maps imposed boundary voltages $g$ to induced boundary currents $\gamma\frac{\partial u}{\partial\vec{n}}|_{\partial\Omega}$, where $\vec{n}$ denotes the unit outward normal. In this study, the DtN operator is approximated using $L = 20$ distinct boundary voltage patterns defined by
$g_l = \cos(2\pi(x\cos(\theta_l) + y\sin(\theta_l)))$,
where $\theta_l = 2\pi l/20$. Operator-based measurements are collected from $m = 128$ uniformly distributed sensors along the four boundaries. This formulation is substantially more challenging than solution-based inversion, since each measurement corresponds to an entire PDE operator rather than discrete solution values at sensor locations. IGNO effectively addresses this complexity through its dual-encoder architecture. The coefficient encoder $E_{\bt_{\bm{\beta}_1}}$ compresses the high-dimensional conductivity field into a compact latent variable $\bm{\beta}_1$, while the boundary encoder $E_{\bt_{\bm{\beta}_2}}$ represents each boundary condition as a one-hot latent vector $\bm{\beta}_2$. Together, these latent representations enable the solution decoder to efficiently predict the full operator response within a unified generative framework. Training is conducted in a purely physics-informed manner by minimizing the residuals of the governing elliptic PDE, eliminating the need for any labeled conductivity–operator pairs. Further implementation details are provided in the Supplementary Methods of the SI, while the performance of IGNO in this problem is reported in Supplementary Table 5.

Fig. \ref{fig:EIT_smh}(b) shows the recovered conductivity fields and pointwise absolute errors obtained by IGNO under different noise levels for an in-distribution test case, where the conductivity follows trigonometric patterns representative of the training distribution (left panel of Fig. \ref{fig:EIT_smh}(a)). IGNO achieves RMSEs of 0.44\%, 1.28\%, and 2.77\% for SNR values of 50, 25, and 15 dB, respectively, accurately capturing both smooth variations and fine-scale features. The reconstruction quality remains high even under severe noise, with errors distributed relatively uniformly across the domain. The framework naturally accommodates all $L=20$ boundary conditions via its dual-encoder structure, efficiently incorporating operator information without requiring repeated PDE solves during training. 
To evaluate generalization beyond the training distribution, we tested an out-of-distribution case in which the conductivity is drawn from a shifted parameter distribution, producing markedly different spatial patterns (right panel of Fig. \ref{fig:EIT_smh}(a)). Despite the substantial distribution shift, IGNO maintains robust performance, with RMSEs of 2.19\%, 2.46\%, and 2.93\% for SNR values of 50, 25, and 15 dB, respectively. While errors are moderately higher than in the in-distribution case, Fig. \ref{fig:EIT_smh}(c) indicates that the framework successfully reconstructs the overall conductivity structure and fine-scale variations. The modest degradation in performance compared to the in-distribution case highlights IGNO’s strong extrapolation capabilities.

These results underscore a key advantage of IGNO: the ability to perform operator-based inversion without retraining or repeated PDE solves for each boundary condition. Once trained, the model can accommodate any combination of boundary inputs by adjusting the boundary latent representation, enabling efficient inversion from operator-valued measurements that would be intractable for existing physics-informed DNOs or PIML-based methods. The dual-encoder design encodes both coefficient and boundary condition information into structured latent representations, providing a unified treatment of diverse measurement modalities. Furthermore, IGNO can handle changes in sensor configurations without retraining, in contrast to data-driven DNOs that require extensive labeled pairs and retraining for new sensor arrangements. These capabilities demonstrate IGNO’s practicality for realistic applications where measurements are noisy, multimodal, and acquired from varying sensor setups, such as in medical imaging and non-destructive testing.
\begin{figure}[tb]
\centering
\includegraphics[width=0.6\textwidth]{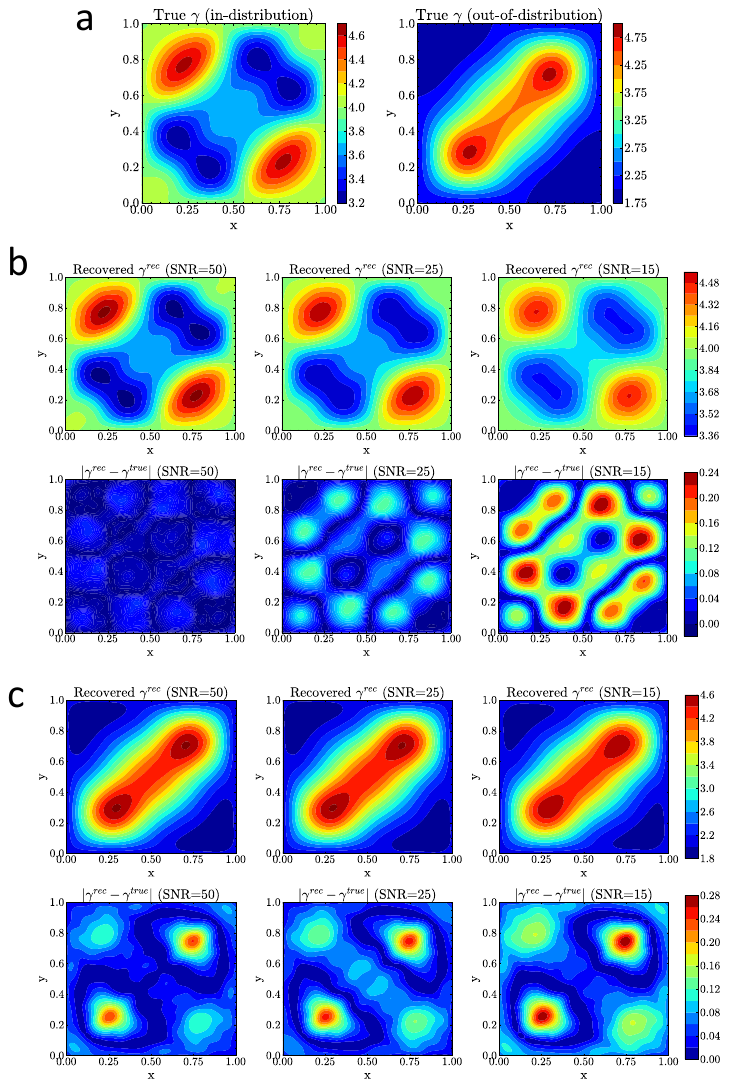}
\caption{Performance of IGNO in solving the EIT problem: (a) the truth conductivity $\gamma$ in In-distribution (left) and Out-of-distribution (right) cases; (b) recovered conductivity $\gamma^{rec}$ and corresponding pointwise absolute errors under different noise levels in In-distribution case; (c) recovered conductivity $\gamma^{rec}$ and corresponding pointwise absolute errors under different noise levels in Out-of-distribution case.}
\label{fig:EIT_smh}
\end{figure}

\subsection{Generalization to out-of-distribution targets}
Beyond the out-of-distribution EIT results presented above, we systematically assessed IGNO’s ability to extrapolate to coefficient distributions that differ substantially from the training data. For continuous Darcy flow problems, we tested on permeability fields with frequency parameters drawn from entirely non-overlapping ranges compared to training, i.e., $(\omega_1,\omega_2) \sim \text{U}(7\pi/4,2\pi)$. For piecewise-constant problems, a different cutoff Gaussian process, $\mathcal{GP}(0,(-\Delta+16I)^{-2})$, was used to generate more complex phase geometries. The performance of both methods in recovering continuous and discontinuous targets under the out-of-distribution case is reported in Supplementary Tables 2 and 4 of the SI, respectively.
\begin{figure}[!htbp]
\centering
\includegraphics[width=0.6\textwidth]{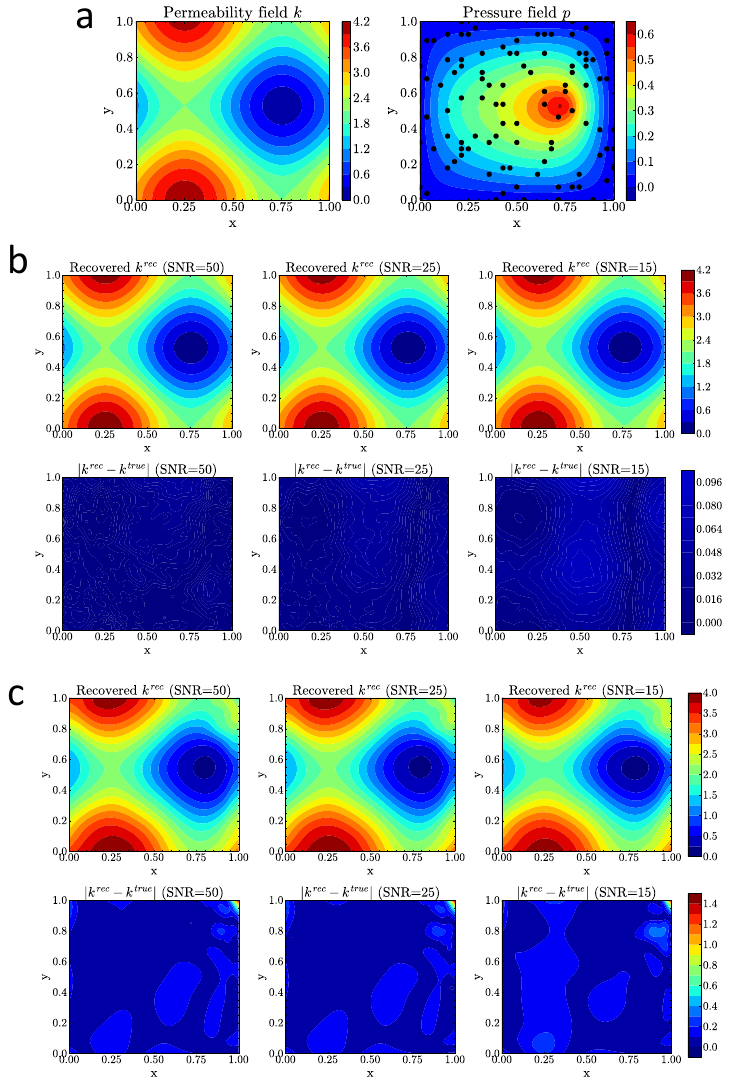}
\caption{Recovery of \textbf{continuous coefficients} with solution-based measurements (\textbf{Out-of-distribution case}): (a) the ground truth permeability field $k$ and the corresponding pressure field $p$ (black dots denote $m=100$ random sensors); (b) recovered permeability $k^{rec}$ and corresponding pointwise absolute errors obtained by IGNO under different noise levels; (c) recovered permeability $k^{rec}$ and corresponding pointwise absolute errors obtained by PI-DIONs under different noise levels.}
\label{fig:smh_out}
\end{figure}

For continuous coefficients, IGNO maintained RMSEs of 0.64\%, 1.13\%, and 1.94\% across noise levels for the out-of-distribution target. This represents only a modest degradation from in-distribution performance and remains 3–5 times more accurate than the PI-DIONs method, which yields RMSEs of 4.76\%, 4.84\%, and 6.39\% for SNR = 50, 25, and 15 dB. As shown in Fig. \ref{fig:smh_out}(b), IGNO successfully reconstructs both large-scale and fine-scale structures, with pointwise errors remaining low even under high noise levels. By contrast, Fig. \ref{fig:smh_out}(c) shows that the PI-DIONs method produces substantial distortions and higher error concentrations, particularly in regions of rapidly varying permeability. Moreover, the RMSE of PI-DIONs increases sharply with noise, highlighting its limited robustness to measurement errors. In contrast, IGNO maintains consistently low errors across all noise levels, demonstrating both robustness and strong generalization to out-of-distribution targets.

\begin{figure}[tb]
\centering
\includegraphics[width=0.6\textwidth]{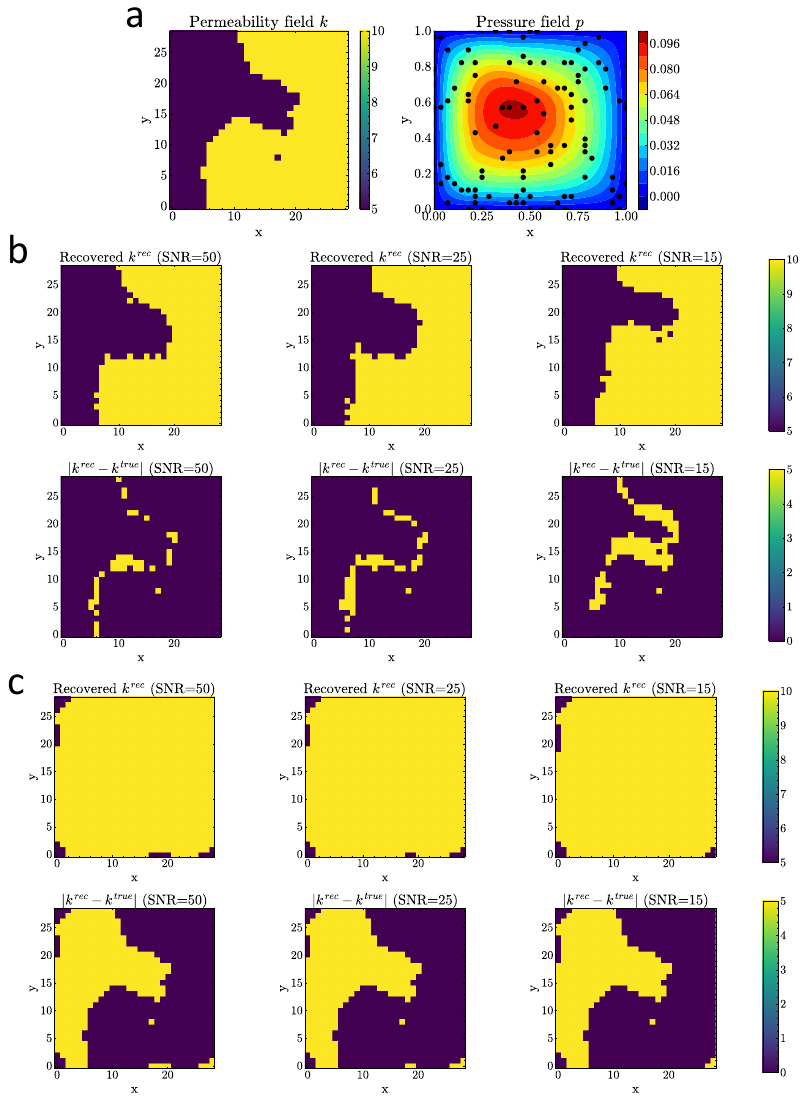}
\caption{Recovery of \textbf{piecewise-constant coefficients} with solution-based measurements (\textbf{Out-of-distribution case}): (a) the true permeability field $k$ (Phase 1 shown in yellow) and the corresponding pressure field $p$ (black dots denote $m=100$ random sensors); (b) recovered permeability $k^{rec}$ and corresponding pointwise absolute errors obtained by IGNO under different noise levels; (c) recovered permeability $k^{rec}$ and corresponding pointwise absolute errors obtained by PI-DIONs under different noise levels.}
\label{fig:pwc_out}
\end{figure}

For piecewise-constant coefficients, IGNO achieves $I_{corr}$ values of 0.960, 0.949, and 0.914 across SNR = 50, 25, and 15 dB, respectively, indicating strong morphological agreement despite the distribution shift. As shown in Fig. \ref{fig:pwc_out}(b), the recovered permeability fields $k^{rec}$ closely resemble the true fields, with pointwise errors concentrated primarily along phase boundaries. By contrast, Fig. \ref{fig:pwc_out}(c) shows that the PI-DIONs method fails consistently across all noise levels, producing $I_{corr}$ values around 0.78 and recovering a single phase almost everywhere, resulting in completely incorrect predictions. These results demonstrate that the latent space learned by IGNO captures fundamental physical structures rather than merely memorizing training data, enabling robust generalization to unseen coefficient patterns.

\section{Discussion}\label{sec:discussion}
The results presented above demonstrate that IGNO provides a robust, accurate, and computationally efficient framework for solving diverse PDE inverse problems. Across all benchmarks, including continuous and piecewise-constant coefficients in Darcy flow and operator-based EIT inversion, IGNO consistently outperforms the state-of-the-art method, achieving reconstruction errors 3 to 6 times lower while maintaining fidelity to fine-scale structures and sharp interfaces. These advances establish IGNO as a broadly applicable methodology that effectively addresses long-standing challenges in inverse problem solving, such as data efficiency, noise robustness, and generalization to unseen coefficient distributions.

The superior performance of IGNO stems from its \textbf{latent-space generative architecture}, which transforms high-dimensional, irregular coefficient recovery into smooth, low-dimensional optimization problems. By embedding both coefficients and boundary conditions into structured latent representations, IGNO produces well-behaved gradients during inversion that align with physically meaningful variations, substantially alleviating the ill-posedness that plagues direct inversion methods. This advantage is particularly evident in discontinuous coefficient recovery, where IGNO accurately reconstructs sharp phase interfaces, achieving cross-correlations above 0.95 even under severe noise, whereas the PI-DIONs method fails due to undefined gradients at discontinuities. The incorporation of PDE residual constraints further stabilizes inversion by enforcing consistency with governing laws, while the normalizing flow provides statistically informed initialization that accelerates convergence and mitigates the risk of poor local minima. This combination of latent modeling, physics-informed constraints, and probabilistic initialization enables IGNO to maintain robust performance even under high measurement noise or out-of-distribution conditions.

The \textbf{physics-informed training paradigm} underpins IGNO’s exceptional data efficiency by eliminating reliance on costly labeled datasets. IGNO requires only coefficient fields and boundary conditions for training, with PDE residuals, formulated in weak or strong form, serving as supervisory signals. This self-supervised approach stands in sharp contrast to purely data-driven DNOs, which typically demand tens of thousands of paired input–output samples generated via expensive forward simulations or physical experiments. Such data efficiency is particularly advantageous in domains like seismic imaging, where acquiring large labeled datasets can be ethically or practically challenging, or in engineering applications where high-fidelity simulations are computationally prohibitive. Moreover, the incorporation of PDE residual constraints not only reduces data requirements but also enforces physical consistency between reconstructed coefficients and corresponding solutions, further enhancing the stability and reliability of the inversion process.

Another key advantage of IGNO is its \textbf{unified treatment of diverse measurement modalities} within a single framework. Unlike existing physics-informed DNOs and PIML-based methods, which are restricted to solution-based observations at fixed sensor locations, IGNO naturally accommodates both solution-based measurements and operator-valued data, such as DtN maps in EIT. Its dual-encoder design separately compresses boundary conditions and coefficients into latent variables, enabling efficient inversion from operator data without repeatedly solving families of PDEs during training or inference. This flexibility not only reduces computational costs but also extends applicability to complex experimental settings where measurements are multimodal, sparse, or noisy. For the EIT problem, IGNO successfully reconstructs conductivity from 20 distinct boundary voltage patterns, achieving RMSEs below 3\% even under high noise, while maintaining robust performance for out-of-distribution targets with spatial patterns markedly different from the training data. This operator-based inversion capability represents a fundamental advance, as no existing physics-informed DNO can handle such measurements.

The results further highlight IGNO’s strong generalization capabilities. For out-of-distribution targets with frequencies or phase geometries not encountered during training, IGNO maintains low reconstruction errors and accurately captures essential physical structures, including phase topology in discontinuous media and conductivity variations in EIT. This demonstrates that IGNO learns a physically meaningful latent space that encodes fundamental structural properties of the coefficient fields, rather than merely memorizing the training data. In contrast, the state-of-the-art method exhibits obvious performance degradation under the same conditions, underscoring the limitations of direct inversion in high-dimensional, unstructured coefficient spaces. In addition, IGNO demonstrates significant computational efficiency, enhancing its practical utility. Once trained, a single model can handle multiple inverse problems with varying observations, sensor configurations, or boundary inputs without retraining. Gradient-based optimization in the low-dimensional latent space converges within 200–500 steps, with each iteration requiring only forward passes through the decoders rather than costly PDE solves. For the largest problem tested, i.e., the EIT problem, individual inversions complete in under 1 minute, representing a substantial reduction in computational cost compared to traditional PDE-constrained optimization, which requires hundreds of forward and adjoint PDE solves per inversion.

Despite these advantages, several limitations and opportunities for future development merit consideration. First, IGNO requires a representative training set of coefficient fields to learn the latent space effectively, and its performance may degrade for coefficient distributions that are extremely far from the training manifold. Similarly, highly anisotropic or multiscale problems with features exceeding the capacity of the chosen latent dimension may pose challenges for accurate reconstruction. Future work could investigate adaptive latent dimensions and hierarchical latent representations to further enhance robustness in highly heterogeneous settings. Second, the current framework focuses on deterministic inversion, producing point estimates of unknown coefficients. Extending IGNO to probabilistic inversion that quantifies uncertainty in recovered parameters would provide valuable information for decision-making in high-stakes applications, such as medical diagnostics or infrastructure assessment. The existing latent space structure and normalizing flow components provide a natural foundation for such extensions via variational inference or sampling-based approaches.

Overall, IGNO establishes a scalable, versatile, and unified framework for PDE inverse problems, bridging physics-informed machine learning, neural operator learning, and generative modeling. By leveraging structured latent representations, enforcing PDE constraints, and performing efficient latent-space optimization, IGNO effectively addresses key challenges in data efficiency, noise robustness, and generalization. Beyond the canonical benchmarks presented here, this methodology is broadly applicable to real-world scientific and engineering problems, including geophysical exploration, medical imaging, nondestructive testing, and materials characterization, where measurements are often noisy, multimodal, and sparsely sampled. The combination of accuracy, efficiency, and flexibility positions IGNO as a powerful tool for advancing computational inverse problem solving in high-dimensional, real-world scenarios.

\section{Methods}\label{sec:methods}
\subsection{Problem formulation}
We consider PDE inverse problems governed by the general PDE system:
\begin{equation}\label{eq:pde}
\begin{array}{ll}
\mathcal{N}(u, a) = s,\quad \text{in}\ \Omega\\
\mathcal{B}(u) = g,\quad \text{on}\ \partial\Omega
\end{array}
\end{equation}
where $u\in\mathcal{U}(\Omega)$ denotes the PDE solution, $a\in\mathcal{A}(\Omega)$ is the unknown coefficient function, $s\in\mathcal{S}(\Omega)$ is the source term, and $g\in\mathcal{G}(\partial\Omega)$ represents boundary conditions on domain $\Omega\in\mathbb{R}^d$ with the Lipschitz boundary $\partial\Omega$. The operators $\mathcal{N}$ and $\mathcal{B}$ denote the differential and boundary operators, respectively, and $\mathcal{U}, \mathcal{A}, \mathcal{S}, \mathcal{G}, \mathcal{H}$ represent suitable function spaces over their respective domains. 

The objective of the inverse problem is to determine unknown quantities, i.e., the coefficient $a$, in the PDE system from (possibly noisy) measurements of the solution $u$. Formally, let $\mathcal{M}(u)$ denote the measurement operator, the form of which depends on the accessibility of sensors:
\begin{itemize}
    \item \textbf{Solution-based measurements.} In this setting, observations consist of solution values at $m$ sensor locations $\bm{X}_m = \{\bm{x}_1, ..., \bm{x}_m\} \subset \Omega$. Therefore, the measurement operator is defined as $\mathcal{M}_{sol} = (u(x_1), ..., u(x_m)) \in \mathbb{R}^m$. The inverse problem then seeks a coefficient $a$ such that the corresponding PDE solution $u$ matches $\mathcal{M}_{sol}$.
    \medskip
    \item \textbf{Operator-based measurements.} When interior measurements are unavailable, one seeks to impose a set of boundary conditions $\{ g_1, g_2, \dots, g_L \}$ on $\partial\Omega$ and record the corresponding measurements $\{ \Lambda_a[g_1], \dots, \Lambda_a[g_L] \}$ on boundary, where $\Lambda_a$ denotes the boundary observation operator. For instance, in EIT, $\Lambda_a$ corresponds to the DtN operator that maps an imposed voltage $g$ to the induced current $a\nabla u\cdot\vec{n}$, with $\vec{n}$ being the unit outward normal vector. The resulting measurement data can be expressed as: 
    \begin{equation}\label{eq:op_obs}
    \mathcal{M}_{\mathrm{op}}: g_l(\bm{x}_i) \mapsto \Lambda_a[g_l](\bm{x}_i),\quad l=1,\cdots,L;\ i=1,\cdots,m.
    \end{equation}
    where $\bm{X}_m=\{\bm{x}_1,\cdots,\bm{x}_m\}\subset\partial\Omega$ denotes boundary sensors. An illustration of the operator-based measurement $\mathcal{M}_{op}$ in the EIT problem is provided in Supplementary Figure 2 of the SI. The inverse problem in this setting seeks to recover the coefficient $a$ such that the induced operator $\Lambda_a$ matches the measured operator $\mathcal{M}_{op}$.
\end{itemize} 
\subsection{The Inverse Generative Neural Operator (IGNO) framework}
\paragraph{Architecture overview.} 
The IGNO framework consists of four neural network components that together enable efficient, physics-informed inversion, as illustrated in Fig.~\ref{fig:model}. Two encoder networks $E_{\bt_{\bm{\beta}_1}}$ and $E_{\bt_{\bm{\beta}_2}}$ compress high-dimensional inputs into low-dimensional latent representations. Specifically, the coefficient encoder $E_{\bt_{\bm{\beta}_1}}$ maps a coefficient field $a$ to a latent variable $\bm{\beta}_1\in\mathbb{R}^{d_{\bm{\beta}_1}}$, while the boundary encoder $E_{\bt_{\bm{\beta}_2}}$ maps a boundary condition $g$ to a latent variable $\bm{\beta}_2\in\mathbb{R}^{d_{\bm{\beta}_2}}$. These are concatenated to form the joint latent representation $\bm{\beta}=(\bm{\beta}_1,\bm{\beta}_2)$. Two decoder networks then map latent representations to outputs: the coefficient decoder $\mathcal{G}_{\bt_a}$ reconstructs the coefficient field $a$ from $\bm{\beta}_1$, while the solution decoder $\mathcal{G}_{\bt_u}$ predicts the PDE solution $u$ from the combined latent representation $\bm{\beta}$. Both decoders are parameterized using the MultiONet architecture \cite{zang2025dgenno}, an enhanced variant of DeepONet that aggregates information across multiple network layers to improve representational capacity. The detailed architecture of the MultiONet is described in Supplementary Note 1 of SI, with a schematic illustration presented in Supplementary Figure 1. For solution-based inverse problems with fixed boundary conditions, the boundary encoder is omitted, and $\bm{\beta}=\bm{\beta}_1$ serves as the sole latent representation. In contrast, for operator-based problems, both encoders are utilized to accommodate varying boundary conditions while maintaining a shared coefficient representation across all boundary inputs.

\paragraph{Physics-informed training.}
All encoder and decoder parameters, $\bt=(\bt_u,\bt_a,,\bt_{\bm{\beta}_1}, \bt_{\bm{\beta}_2})$, are jointly optimized using a physics-informed loss function that embeds the governing PDEs directly into training, eliminating the need for labeled solution data:
\begin{equation}\label{eq:loss}
\mathcal{L}(\bt) = \lambda_{pde} \mathcal{L}_{pde}(\bt) + \lambda_{bd} \mathcal{L}_{bd}(\bt_u,\bt_{\bm{\beta}_1},\bt_{\bm{\beta}_2}) + \lambda_{rec} \mathcal{L}_{rec}(\bt_a,\bt_{\bm{\beta}_1},\bt_{\bm{\beta}_2}),
\end{equation}
where $\lambda_{pde}$, $\lambda_{bd}$, and $\lambda_{rec}$ are positive weights balancing each term. Given training data $\mathcal{D} = \{g^{(n)},a^{(n)}\}_{n=1}^N$, which contains only coefficient fields and boundary conditions, the \textbf{PDE loss} $\mathcal{L}_{pde}$ enforces that the predicted coefficients and solutions satisfy the governing equations through residual minimization:
\begin{equation}\label{eq:loss_pde}
\mathcal{L}_{pde} = \frac{1}{N} \sum_{n=1}^N \| \mathcal{R}^{(n)}(\mathcal{G}_{\bt_u}(\bm{\beta}^{(n)}),\mathcal{G}_{\bt_a}(\bm{\beta}_1^{(n)})) \|_2^2,
\end{equation}
where $\mathcal{R}^{(n)} = (r^{(n)}_{w_1}, \dots, r^{(n)}_{w_K})\in\mathbb{R}^K$ is a vector of weighted residuals, with each component defined as:
\begin{equation}\label{eq:residual}
r^{(n)}_{w_k} = \int_\Omega
\big[ \mathcal{N}(\mathcal{G}_{\bt_u}(\bm{\beta}^{(n)}),\mathcal{G}_{\bt_a}(\bm{\beta}_1^{(n)})) - s^{(n)} \big] w_k \, d\bm{x}.
\end{equation}
Here, $w_k$ denotes the weighting function. For strong-form residuals, $w_k$ is chosen as a Dirac delta function $\delta(\bm{x} - \bm{x}_k)$, yielding pointwise residual evaluation at collocation points $\bm{x}_k$. For weak-form residuals, preferred when dealing with discontinuous coefficients or reduced solution regularity, compactly supported radial basis functions (CSRBFs)~\cite{zang2023particlewnn,wendland1995piecewise} are employed as weighting functions. These localized functions significantly reduce computational cost while improving handling of singularities and adaptation to complex geometries.

The \textbf{boundary loss}, $\mathcal{L}_{bd}$, enforces boundary conditions:
\begin{equation}\label{eq:loss_bd}
\mathcal{L}_{bd} = \frac{1}{N}\sum_{n=1}^N \left\| \mathcal{B}^{(n)}\big(\mathcal{G}_{\bt_u}(\bm{\beta}^{(n)})(\Xi_{bd})\big) - \bm{g}^{(n)} \right\|_2^2,
\end{equation}
where $\bm{g}^{(n)} = g^{(n)}(\Xi_{bd}) \in \mathbb{R}^{N_{bd}}$ represents boundary values at the boundary sensor set $\Xi_{bd} = (\xi_{bd,1}, \ldots, \xi_{bd,N_{bd}}) \subset \partial\Omega$.

The \textbf{reconstruction loss}, $\mathcal{L}_{rec}$, ensures that the coefficient decoder accurately reconstructs input coefficients from their latent representations:
\begin{equation}\label{eq:loss_rec}
\mathcal{L}_{rec} = \frac{1}{N} \sum_{n=1}^N \left\| \mathcal{G}_{\bt_a}(\bm{\beta}_1^{(n)})(\Xi_a) - \bm{a}^{(n)} \right\|_2^2,
\end{equation}
where $\bm{a}^{(n)}\in \mathbb{R}^{N_a}$ denotes coefficient values sampled at grid points $\Xi_a = (\xi_{a,1}, \ldots, \xi_{a,N_a}) \subset \Omega$. For piecewise-constant coefficients, this loss is replaced by a cross-entropy formulation in which the decoder predicts phase membership probabilities rather than permeability values, preventing gradient instability at discontinuities. Further methodological details are provided in the Supplementary Methods of the SI.

The model is trained using stochastic gradient descent with the ADAM optimizer, an initial learning rate of $1\times10^{-3}$ (halved every 2,500 epochs), a batch size of 50, and a total of 10,000 epochs until convergence. This purely physics-driven training requires no precomputed PDE solutions, in contrast to data-driven DNOs that depend on large labeled datasets. Once trained, the solution decoder $\mathcal{G}_{\bt_u}$ serves as a fast, surrogate PDE solver, while the coefficient decoder $\mathcal{G}_{\bt_a}$ enables efficient coefficient reconstruction from optimized latent representations. The algorithm for training IGNO is provided in Supplementary Note 3 of the SI.

\subsection{Gradient-based inversion in latent space}
With trained encoders and decoders, inverse problems are solved via gradient-based optimization directly in the low-dimensional, structured latent space, rather than in the original high-dimensional and often irregular coefficient space. This transformation substantially enhances both tractability and stability of the inversion process.

\paragraph{Solution-based case.} 
For inverse problems with solution-based measurements $\mathcal{M}_{sol}$, where the boundary condition is fixed, the latent variable reduces to $\bm{\beta} = \bm{\beta}_1$. The inversion is formulated as the following optimization problem:
\begin{equation}\label{eq:obj_solution}
\begin{aligned}
\bm{\beta}_1^* &= \arg\min_{\bm{\beta}_1} \mathcal{F}(\bm{\beta}_1) := \mathcal{F}_{data}(\bm{\beta}_1) + \mathcal{F}_{pde}(\bm{\beta}_1) \\
&= \frac{\rho_{data}}{m} \sum_{i=1}^m \left| \mathcal{G}_{\bt^*_u}(\bm{\beta}_1)(\bm{x}_i) - u(\bm{x}_i) \right|^2 + \rho_{pde} \left\| \mathcal{R}\left(\mathcal{G}_{\bt_u^*}(\bm{\beta}), \mathcal{G}_{\bt_a^*}(\bm{\beta}_1) \right) \right\|_2^2.
\end{aligned}
\end{equation}
Here, the term $\mathcal{F}_{data}$ measures the data mismatch between predicted and observed solution at sensor locations $\bm{X}_m$, weighted by $\rho_{data}$. The term $\mathcal{F}_{pde}$ serves as a physics-based regularization term, weighted by $\rho_{pde}$, enforcing consistency with the PDE by minimizing the residual norm.

\paragraph{Operator-based case.} 
For inverse problems with operator-based measurements $\mathcal{M}_{op}$ as in \eqref{eq:op_obs}, the full latent variable $\bm{\beta}=(\bm{\beta}_1,\bm{\beta}_2)$ is employed, but optimization is performed only with respect to $\bm{\beta}_1$ to recover the unknown coefficient field:
\begin{equation}\label{eq:obj_operator}
\begin{aligned}
\bm{\beta}_1^* &= \arg\min_{\bm{\beta}_1} \mathcal{F}(\bm{\beta}_1) := \mathcal{F}_{data}(\bm{\beta}_1) + \mathcal{F}_{pde}(\bm{\beta}_1) \\
&= \frac{\rho_{data}}{mL} \sum_{l=1}^{L} \sum_{i=1}^m \left| \Lambda_{\mathcal{G}_{\bt_a^*}(\bm{\beta}_1)}[g_l](\bm{x}_i) - \Lambda_a[g_l](\bm{x}_i) \right|^2  + \frac{\rho_{pde}}{L} \sum_{l=1}^{L} \left\| \mathcal{R}^{(l)}\left( \mathcal{G}_{\bt_u^*}(\bm{\beta}^{(l)}), \mathcal{G}_{\bt_a^*}(\bm{\beta}_1) \right) \right\|_2^2.
\end{aligned}
\end{equation}
In this case, the term $\mathcal{F}_{data}$ quantifies the mismatch between predicted and true operator outputs across all $L$ boundary conditions and $m$ sensors, weighted by $\rho_{data}$. The term $\mathcal{F}_{pde}$  enforces PDE consistency for each boundary condition, with $\bm{\beta}^{(l)}=(\bm{\beta}_1, E_{\bt^*_{\bm{\beta}_2}}(g_l))$ combining the coefficient latent representation and the encoded boundary condition.

The latent-space optimization is performed using the ADAM optimizer with an initial learning rate of 0.01, decayed by a factor of $2/3$ every 250 steps for solution-based cases, or adaptively per problem for operator-based cases. Convergence typically occurs within 200–500 iterations. Gradients with respect to $\bm{\beta}_1$ are computed efficiently via automatic differentiation through the trained decoders. After convergence, the optimal latent vector $\bm{\beta}^*_1$ is passed through the coefficient decoder $\mathcal{G}_{\bt^*_a}$ to reconstruct the recovered coefficient field $a$. The algorithm for latent-space optimization is provided in Supplementary Note 3 of the SI.

\subsection{Normalizing flow–based a priori initialization}
To enhance the efficiency and robustness of latent-space optimization, we introduce a normalizing flow model $F_{\bt_{NF}}:\mathbb{R}^{\bm{\beta}_1}\rightarrow\mathbb{R}^{\bm{\beta}_1}$ that maps the learned latent space of $\bm{\beta}_1$ to a standard multivariate normal distribution $\mathcal{N}(\mathbf{0},\mathbf{I})$. Specifically, we employ the Real-valued Non-Volume Preserving (RealNVP) architecture \cite{dinh2016density}, an invertible neural network that enables tractable density estimation via the change-of-variables:
\begin{equation}
\log p_{\bm{\beta}_1}(\bm{\beta}_1) = \log p_{\mathbf{z}}(\bm{z}) + \log\left|\det \frac{\partial\bm{z}}{\partial \bm{\beta}_1}\right|,
\end{equation}
where $\bm{z} = F_{\bt_{NF}}(\bm{\beta}_1)$ follows the standard Gaussian density $p_{\mathbf{z}}$. A detailed description of the flow model is provided in Note 2 of the SI. The flow is trained by minimizing the negative log-likelihood over the training set of learned latent representations obtained from the trained IGNO model:
\begin{equation}
\mathcal{L}_{NF}(\bt_{NF}) = -\frac{1}{N}\sum_{n=1}^N \left[ \log p_{\mathbf{z}}\big(F_{\bt_{NF}}(\bm{\beta}^{(n)}_1)\big) + \log\left|\det \frac{\partial F_{\bt_{NF}}(\bm{\beta}_1^{(n)})}{\partial \bm{\beta}_1}\right| \right].
\end{equation}

Once trained, the flow model provides statistically informed initialization by sampling $\bm{z} \sim \mathcal{N}(\mathbf{0}, \mathbf{I})$ from the standard Gaussian and mapping to the latent space via the inverse flow $\bm{\beta}_1^{init} = F_{\bt^*_{NF}}^{-1}(\bm{z})$. This initialization strategy captures the distribution of physically plausible coefficients learned during training, substantially accelerating convergence and improving robustness compared to random initialization, particularly in noisy or ill-posed scenarios.

\subsection{Evaluation metrics}
Two quantitative metrics are employed to assess reconstruction performance. For continuous coefficients, the \textbf{relative root-mean-square error (RMSE)} measures reconstruction accuracy as:
\begin{equation}
\text{RMSE} = \sqrt{\frac{\sum_{i} (a^{rec}(\bm{x}_i)-a^{true}(\bm{x}_i))^2}{\sum_{i} (a^{true}(\bm{x}_i))^2}},
\end{equation}
where $a^{rec}$ and $a^{true}$ denote recovered and ground-truth coefficients evaluated at grid points.

For discontinuous targets, morphological similarity between piecewise-constant fields is quantified using the \textbf{cross-correlation indicator ($I_{corr}$)} \cite{bourke1996cross}:
\begin{equation}
I_{corr} = \frac{\sum_i (\tilde{a}^{true}(\bm{x}_i))^2(\tilde{a}^{rec}(\bm{x}_i))^2}{\sqrt{\sum_i (\tilde{a}^{ture}(\bm{x}_i))^2}\sqrt{\sum_i (\tilde{a}^{rec}(\bm{x}_i))^2}},
\end{equation}
where $\tilde{a}$ denotes coefficients rescaled to [0,1]. The indicator $I_{corr}$ ranges from 0 to 1, with values close to 1 indicating strong morphological agreement between the reconstructed and true phase topologies.

\section{Code Availability Statement}
All code necessary to reproduce the results presented in this paper will be made publicly available upon publication in our GitHub repository.

\section{Data Availability Statement}
Upon publication, the training datasets and pre-trained models will be made publicly available via Kaggle dataset. The raw data to create the figures of the main text can be made available from the corresponding author upon reasonable request.

\section*{Acknowledgement}
The work is partially supported by National Natural Science Foundation of China (U21A20425) and a Key Laboratory of Zhejiang Province.

\newpage
\bibliographystyle{elsarticle-num}
\bibliography{ref.bib}

\clearpage
\appendix
\setcounter{figure}{0}
\setcounter{table}{0}
\clearpage
\begin{center}
    {\LARGE \bfseries Supplementary Material}
\end{center}
\vspace{1em}
\section{Supplementary Note 1: The MultiONet Architecture}\label{sec:multionet}
The MultiONet architecture extends the standard DeepONet by aggregating information from multiple intermediate layers of both branch and trunk networks, enriching the learned representation without increasing trainable parameters. The architecture, illustrated in Fig. \ref{fig:MultiONet}, comprises two subnetworks:
\begin{itemize}
    \item \textbf{Trunk network}: Encodes the spatial coordinates $\bm{x} \in \Omega$ of the output field.
	\item \textbf{Branch network}: Encodes a latent vector $\bm{\beta}$ representing a compact descriptor of the input functions.
\end{itemize}
Unlike the original DeepONet, MultiONet computes a weighted average of inner products between layer-wise features from multiple network depths rather than relying solely on last-layer features. Formally, the MultiONet mapping $\mathcal{G}(\bm{\beta})(\bm{x})$ is expressed as:
\begin{equation}\label{eq:multionet}
\mathcal{G}(\bm{\beta})(\bm{x}) =
\frac{1}{l} \sum_{k=1}^{l} w^{(k)}
\left( b^{(k)}(\bm{\beta}) \odot t^{(k)}(\bm{x}) \right) + b_0,
\end{equation}
where $b^{(k)}(\bm{\beta})$ and $t^{(k)}(\bm{x})$ denote the outputs from the $k$-th layers of the branch and trunk networks, $l$ represents the total number of layers, $w^{(k)}$ indicates trainable weights, $b_0$ is the bias term, and $\odot$ represents the inner product operation.

\section{Supplementary Note 2: Normalizing Flow for Initialization}\label{sec:nf_details}
In gradient-based inversion, initial guess quality for the latent variable affects both convergence speed and optimization quality. Random initialization (e.g., from uniform or Gaussian distributions) ignores the trained latent space structure and may lead to inefficient optimization, particularly for highly nonlinear PDE inverse problems. We employ a normalizing flow (NF) model to learn a bijective mapping between the latent space and a standard multivariate normal distribution, enabling informed and probabilistically meaningful initialization. Specifically, we use the RealNVP architecture \cite{dinh2016density}, which consists of invertible coupling layers. Each coupling layer splits the input $\bm{x}\in\mathbb{R}^d$ into $\bm{z}_1\in\mathbb{R}^{d/2}$ and $\bm{z}_z\in\mathbb{R}^{d/2}$, then applies:
\begin{itemize}
    \item Forward: $\bm{y}_1=\bm{z}_1$, $\bm{y}_2=\bm{z}_2\odot \exp(s(\bm{z}_1))+t(\bm{z}_1)$ 
    \item Inverse: $\bm{z}_1=\bm{y}_1$, $\bm{z}_2=(\bm{y}_2-t(\bm{y}_1)) \odot \exp(-s(\bm{y}_1))$ 
\end{itemize}
where $s(\cdot)$ and $t(\cdot)$ are scale and translation networks parameterized by fully connected neural networks (FCNNs).

\section{Supplementary Note 3: Training and Inversion Algorithms}\label{sec:algorithms}
\begin{algorithm}[H]
\caption{Training the IGNO Framework}
\label{alg:genno_training}
\begin{algorithmic}
\Inputs{Training data $\mathcal{D} = \{a^{(n)}, g^{(n)}\}_{n=1}^N$; loss weights $\lambda_{pde}$, $\lambda_{bd}$, $\lambda_{rec}$.}
\Initialize{Model parameters $\bt=(\bt_u, \bt_a, \bt_{\bm{\beta}_1}, \bt_{\bm{\beta}_2})$; learning rate $lr$.}
\While{Convergence or maximum number of iterations not reached}
\State{Obtain latent representations: \begin{equation*}
\bm{\beta}^{(n)}_1 \gets E_{\bm{\beta}_1}(\bm{a}^{(n)}),\quad\bm{\beta}^{(n)}_2 \gets E_{\bm{\beta}_2}(\bm{g}^{(n)}).\quad n=1,\cdots,N
\end{equation*}}
\State{Obtain the predicted PDE solutions and recovered coefficients:
\begin{equation*}
u^{(n)}_{\mathrm{pred}} \gets \mathcal{G}_{\bt_u}(\bm{\beta}^{(n)}),\quad 
a^{(n)}_{\mathrm{pred}} \gets \mathcal{G}_{\bt_a}(\bm{\beta}^{(n)}_1).\quad n=1,\cdots,N 
\end{equation*}}
\State{Compute the loss $\mathcal{L}(\bt)$ and update model parameters with SGD:
$$\bt \leftarrow \bt - lr \odot \nabla_{\bt} \mathcal{L}(\bt).$$}
\If{at every 2500th epoch}
\State {Update the learning rate to $lr \leftarrow lr/2$.}
\EndIf
\EndWhile
\Outputs{Trained encoders $E_{\bt^*_{\beta_1}}$, $E_{\bt^*_{\beta_2}}$, and decoders $\mathcal{G}_{\bt^*_u}$, $\mathcal{G}_{\bt^*_a}$.}
\end{algorithmic}
\end{algorithm}

\begin{algorithm}[H]
\caption{Inversion via Latent Optimization}
\label{alg:inversion}
\begin{algorithmic}[1]
\Inputs{Trained networks $E_{\bt^*_{\bm{\beta}_1}}$, $E_{\bt^*_{\bm{\beta}_2}}$, $\mathcal{G}_{\bt^*_u}$, $\mathcal{G}_{\bt^*_a}$; Trained normalizing flow $F_{\bt^*_{NF}}$; measurements $\mathcal{M}$}
\Initialize{Sample $\bm{z}\sim\mathcal{N}(\bm{0},\bm{I})$; Set $\bm{\beta}_1\leftarrow F_{\bt^*_{NF}}^{-1}(\bm{z})$; learning rate $lr$}
\While{Convergence or maximum number of iterations not reached}
\State{Compute optimization objective $\mathcal{F}(\bm{\beta}_1)$.}
\State{Update $\bm{\beta}_1$ with gradient descent: $$\bm{\beta}_1 \leftarrow \bm{\beta}_1 - lr\cdot\nabla_{\bm{\beta}_1} \mathcal{F}(\bm{\beta}_1)$$}
\If{at every 250th epoch}
\State {Update the learning rate to $lr \leftarrow lr/2$.}
\EndIf
\EndWhile
\Outputs{Recovered coefficient $a^{rec}=\mathcal{G}_{\bt^*_a}(\bm{\beta}_1)$.}
\end{algorithmic}
\end{algorithm}

\section{Supplementary Note 4: Baseline Method Details}
We compare IGNO against physics-informed deep inverse operator networks (PI-DIONs) \cite{cho2024physics}, a recent state-of-the-art method for PDE inverse problems with solution-based measurements. This method employs the DeepONet architecture to parameterize both the unknown coefficient $a$ and the PDE solution $u$. Coordinates $\bm{x}$ are treated as inputs to the trunk networks, while solution-based measurements $\mathcal{M}_{sol}$ are input to the branch networks. The two resulting models are denoted $\mathcal{G}_{\bt_u}$ and $\mathcal{G}_{\bt_a}$, parameterized by $\bt_u$ and $\bt_a$, respectively. Predictions at location $\bm{x}$ are given by $\mathcal{G}_{\bt_u}(\mathcal{M}_{sol})(\bm{x})$ for the solution and $\mathcal{G}_{\bt_a}(\mathcal{M}_{sol})(\bm{x})$ for the coefficient.

Training of PI-DIONs relies on a physics-informed composite loss that combines (i) a physics loss, consisting of strong-form PDE residuals ($\mathcal{L}_{pde}$) and boundary mismatches ($\mathcal{L}_{bd}$), and (ii) a data loss ($\mathcal{L}_{data}$) that penalizes deviations between predicted and observed solutions at fixed sensors. For $N$ training samples with solution-based measurements $\{\mathcal{M}^{(n)}_{sol}\}_{n=1}^N$ collected on sensors $\bm{X}_m=(\bm{x}_1,\dots,\bm{x}_m)\subset\Omega$, the total loss is defined as:
\begin{equation}\label{eq:loss_pidion}
\begin{aligned}
    \mathcal{L} &= \lambda_{physics}(\mathcal{L}_{pde}+\mathcal{L}_{bd})+ \lambda_{data}\mathcal{L}_{data} \\
    & = \frac{\lambda_{physics}}{N}\sum^N_{n=1}\Big( \|\bm{\mathcal{R}}^{(n)}(\mathcal{G}_{\bt_u}(\mathcal{M}^{(n)}_{sol}),\mathcal{G}_{\bt_a}(\mathcal{M}^{(n)}_{sol}))\|^2_2 + \|\bm{\mathcal{B}}^{(n)}(\mathcal{G}_{\bt_u}(\mathcal{M}^{(n)}_{sol})(\Xi_{bd}))-\bm{g}^{(n)}\|^2_2\Big) \\
    &\quad + \frac{\lambda_{data}}{N}\sum^N_{n=1}\|\mathcal{G}_{\bt_u}(\mathcal{M}^{(n)}_{sol})(\bm{X}_m)-\mathcal{M}^{(n)}_{sol}\|^2_2,
\end{aligned}
\end{equation}
where $\bm{\mathcal{R}}^{(n)}=(r^{(n)}_{w_1},\dots,r^{(n)}_{w_K})\in\mathbb{R}^{K}$ denotes the strong-form residual vector. The terms $\lambda_{physics}$ and $\lambda_{data}$ are loss weights.  
Once trained, the recovered coefficient $a$ and solution $u$ for test measurements $\mathcal{M}^{test}_{sol}$ are directly obtained via $\mathcal{G}_{\bt^*_a}(\mathcal{M}^{test}_{sol})$ and $\mathcal{G}_{\bt^*_u}(\mathcal{M}^{test}_{sol})$, respectively. PI-DIONs is not applicable to operator-based inverse problems and serves as a baseline only for solution-based cases.

\section{Supplementary Methods}
\subsection{General Implementation Details}
Unless otherwise specified, all models are trained using the ADAM optimizer with an initial learning rate of $1\times 10^{-3}$, reduced by half every $2500$ epochs. The batch size is set to $50$, and training proceeds for $10,000$ epochs to ensure convergence. For the proposed IGNO framework, loss weights are set as $\lambda_{\mathrm{pde}} = 0.25$, $\lambda_{bd} = 0.5$, and $\lambda_{rec} = 2$. For PI-DIONs, loss weights are set as $\lambda_{physics}=1$ and $\lambda_{data}=2$ for best performance. All experiments are conducted under identical hardware conditions on a 64-core AMD Ryzen CPU with an NVIDIA RTX 4090 GPU.

\subsection{Continuous Coefficient Recovery with Solution-based Measurements}
\subsubsection{Problem Setup}
The governing PDE is the following Darcy’s flow equation:
\begin{equation}
\begin{aligned}
-\nabla(k(x,y)\nabla p(x,y)) &= f(x,y), \quad \text{in} \ \Omega=[0,1]^2, \\
p(x,y) &= 0,\quad \text{in}\ \partial\Omega,
\end{aligned}
\end{equation}
where $k$ is the permeability field, $p$ is the pressure field, and $f=10$ is the source term. Measurements are collected at $m=100$ sensor locations $\bm{X}_m=(\bmx_1,\ldots,\bmx_m)$, with additive Gaussian noise contamination:
\begin{equation}
\mathcal{M}_{sol} = \left(\hat{p}(\bmx_1), \ldots, \hat{p}(\bmx_m)\right), \ \text{where}\ \ \hat{p}(\bmx_i) = p(\bmx_i) + \epsilon_i,\quad \epsilon_i \sim \mathcal{N}(0,\sigma^2).
\end{equation}
Noise levels are characterized by the signal-to-noise ratio (SNR), which is defined as:
\begin{equation}
\text{SNR} = 10\log{10}\left(\frac{\frac{1}{m}\sum_{i=1}^m p^2(\bmx_i)}{\sigma^2}\right).
\end{equation}
In particular, we examine three scenarios: low noise (SNR=50), medium noise (SNR=25), and high noise (SNR=15).

For training IGNO, $N=1000$ synthetic permeability fields are generated as $k(x,y) = 2.1 + \sin(\omega_1 x) + \cos(\omega_2 y)$ with $\omega_1,\omega_2$ sampled independently from the uniform distribution $\text{U}(0, 7\pi/4)^2$. Since the boundary condition is fixed (i.e., $g=0$) across all samples, only the permeability encoder $E_{\bm{\beta}_1}$ is required.
For training PI-DIONs, pressure fields $p^{(n)}$ are computed by solving~\eqref{eq:darcy_smh} via the finite element method (FEM) for each permeability sample $k^{(n)}$. Then, training measurements $\mathcal{M}^{(n)}_{sol}$ are formulated by sampling the pressure values at the same sensor locations $\bm{X}_m$ without added noise. 

To evaluate the performance of both methods on this problem, we consider two classes of test targets:
\begin{itemize}
    \item\textbf{In-distribution target:} permeability fields $k$ generated with $(\omega_1,\omega_2) \sim \text{U}(0,7\pi/4)$, consistent with the training distribution.
    \item\textbf{Out-of-distribution target:} permeability fields $k$ generated with $(\omega_1,\omega_2) \sim \text{U}(7\pi/4,2\pi)$, which lie entirely outside the training distribution.
\end{itemize} 
In both cases, noisy test measurements $\mathcal{M}^{test}_{sol}$ are generated by solving~\eqref{eq:darcy_smh} with FEM and perturbing the pressure values with Gaussian noise at the specified SNR levels.

\subsubsection{Model Setup}\label{sec:model_darcy_smh}
To extract latent representations from the input coefficients with the encoder $E_{\bt_{\bm{\beta}_1}}$, we set $\Xi_a = (\xi_{a,1}, \ldots, \xi_{a,N_a}) \subset \Omega$ as a $29\times 29$ uniform grid, so that the encoder input is $\bm{a}=a(\Xi_a)$. Since the boundary condition is fixed ($g=0$) across all samples, no boundary input is needed (i.e., the boundary encoder vanishes). Moreover, a mollifier $f(x,y)=\sin(\pi x)\sin(\pi y)$ is applied to the output of the solution decoder, i.e.,
\begin{equation}
u_{pred}(x,y) = \mathcal{G}_{\bt_u}(\bm{\beta})(x,y)\cdot f(x,y),
\end{equation}
which enforces the boundary condition automatically and removes the need for an explicit boundary loss during training. Below, we provide details of the network architectures for both methods:
\paragraph{The IGNO method}
\begin{itemize}
    \item \textbf{Encoder $E_{\bt_{\bm{\beta}_1}}$.} It consists of a Convolutional Neural Network (CNN) followed by a Feed-Forward Fully Connected Network (FFCN). The CNN has three hidden layers with 64 output channels each, kernel size $(3,3)$, and stride 2. The FFCN has two hidden layers of 128 neurons each. The SiLU activation is applied to all hidden layers, while the output layer uses Tanh to constrain the latent variables to a bounded cubic region.
    \item \textbf{Solution Decoder $\mathcal{G}_{\bt_u}$.} The solution decoder $\mathcal{G}_{\bt_u}$ adopts the MultiONet architecture. Both branch and trunk networks are FFCNs with 6 hidden layers of 100 neurons each. A custom activation function, Tanh\_Sin, is used in all hidden layers:
    \begin{equation}\label{eq:tanh_sin}
    \text{Tanh\_Sin}(x) = \tanh(\sin(\pi x + \pi)) + x.
    \end{equation}
    \item \textbf{Coefficient Decoder $\mathcal{G}_{\bt_a}$.} The coefficient decoder $\mathcal{G}_{\bt_a}$ adopts the same architecture as $\mathcal{G}_{\bt_u}$.
    \item \textbf{NF model $F_{\bt_{NF}}$.} The NF model consists of three flow steps, each parameterized by a fully connected network with two hidden layers of 64 neurons per layer and SiLU activations.
\end{itemize}
\paragraph{The PI-DIONs method}
\begin{itemize}
    \item \textbf{Decoder $\mathcal{G}_{\bt_u}$.} For fair comparison, the PI-DIONs solution decoder adopts the same architecture as in IGNO. However, the input to the branch network is the solution-based measurement $\mathcal{M}_{sol}$, instead of the latent variable $\bm{\beta}$.
    \item \textbf{Decoder $\mathcal{G}_{\bt_a}$.} Similarly, the coefficient decoder $\mathcal{G}_{\bt_a}$ uses the same architecture as in IGNO, with the branch network taking $\mathcal{M}_{sol}$ as input instead of $\bm{\beta}_1$.
\end{itemize}
\subsubsection{Latent-space Optimization Details}
Gradients of the objective $\mathcal{F}$ with respect to $\bm{\beta}_1$ are computed via PyTorch’s automatic differentiation. The weights $\rho_{data}$ and $\rho_{pde}$ are set to $50$ and $1$, respectively. Optimization is performed using ADAM with an initial learning rate of $0.01$, decayed by a factor of $2/3$ every 250 steps, for a total of 500 updates.

\subsection{Piecewise-Constant Coefficient Recovery with Solution-based Measurements}
\subsubsection{Problem Setup}
The governing equation remains Darcy’s law \eqref{eq:darcy_smh} but with piecewise-constant coefficient:
\begin{equation}
k(x,y) =
\begin{cases}
    10, \quad (x,y)\in \Omega_1\\
    5, \quad (x,y)\in \Omega_2
\end{cases}
\end{equation}
where $\Omega_1$ and $\Omega_2$ represent two disjoint phases (i.e., phase 1 and phase 2) such that $\Omega_1 \cup \Omega_2 = \Omega$. Measurements $\mathcal{M}_{sol}$ are collected on $m=100$ fixed sensors $\bm{X}_m$ randomly sampled in $\Omega$. The measurements are contaminated with Gaussian noise, and three levels of noise are considered: low (SNR=50), medium (SNR=20), and high (SNR=15).
The discontinuous nature of the coefficient field makes this problem particularly difficult for existing methods: (i) Gradient-based methods require differentiability of $k$, but in this case, $k$ is discrete-valued, so gradients with respect to $k$ are not well defined; (ii) Non-gradient methods such as evolutionary algorithms or MCMC may, in principle, handle discrete parameters, but they become computationally prohibitive when $k$ lies in a high- or infinite-dimensional function space, restricting their applicability to only low-dimensional problems. The proposed IGNO framework naturally overcomes these challenges by introducing a low-dimensional and well-structured latent space representation of the original high-dimensional and discontinuous coefficient field. This transforms the otherwise intractable inverse problem into a tractable optimization problem in a continuous latent space, making it both efficient and robust.

To train IGNO, piecewise-constant permeability fields are generated by using a cutoff Gaussian Process $\mathcal{GP}(0,(-\Delta + 9I)^{-2})$ \cite{li2020fourier,zang2025dgenno}. For each GP realization, $k(x,y) = 10$ if the underlying GP-value is greater than $0$ and $k(x,y) = 5$ otherwise. Again, only the permeability encoder $E_{\bm{\beta}_1}$ is required as the boundary condition is fixed across all samples.
For training PI-DIONs, the pressure field $p^{(n)}$ corresponding to each permeability sample $k^{(n)}$ is obtained using FEM. Noise-free pressure values on the fixed sensors $\bm{X}_m$ are then used as training measurements $\mathcal{M}^{(n)}_{sol}$. 

We evaluate both methods on two types of inverse targets:
\begin{itemize}
    \item\textbf{In-distribution target:} The target permeability $k$ is generated from the same distribution used in training, i.e., the cutoff $\mathcal{GP}(0,(-\Delta+9I)^{-2})$.
    \item\textbf{Out-of-distribution target:} The target $k$ is generated from a different distribution, namely the cutoff $\mathcal{GP}(0,(-\Delta+16I)^{-2})$. The higher-order correlation structure of $k$ differs significantly from the training distribution, resulting in more complex geometries of the permeability field.
\end{itemize} 
In both cases, test measurements $\mathcal{M}^{test}_{sol}$ are obtained by solving Darcy’s equation with FEM, and then corrupted with Gaussian noise corresponding to the specified SNR levels.

\subsubsection{Model Setup}\label{sec:model_darcy_pwc}
The coefficient $k$ is represented by an image of size $29\times 29$, denoted by $\Xi_a = (\xi_{a,1}, \ldots, \xi_{a,N_a})$. Each pixel value corresponds to the permeability of the associated phase (either $10$ or $5$). To improve the performance of IGNO in this problem, the coefficient decoder is not trained to predict the coefficient values directly. Instead, it predicts the probability that a spatial location $\bm{x}$ belongs to phase 1. Accordingly, the recovery loss $\mathcal{L}_{rec}$ is replaced by a cross-entropy loss:
\be
\mathcal{L}_{rec}
= \frac{1}{N} \sum_{n=1}^N\sum_{i=1}^{N_a} \left[ z^{(n)}_i \log \sigma\big(\mathcal{G}_{\bm{\theta}_a}(\bm{\beta}_1^{(n)})(\xi_{a,i}) \big) + (1-z^{(n)}_i) \log\big(1 - \sigma(\mathcal{G}_{\bm{\theta}_a}(\bm{\beta}_1^{(n)})(\xi_{a,i}) \big) \right],
\ee
where $\sigma(\cdot)$ is the sigmoid function, and $z_i$ is a binary label indicating the true phase at location $\xi_{a,i}$ ($z_i=1$ for phase 1 and $z_i=0$ for phase 2).
At inference time, the recovered coefficient can be obtained either by sampling from the predicted probability field or, as done in this work, by applying a cut-off rule: values greater than $0.5$ are assigned to phase 1 and others to phase 2. This strategy is not applicable to PI-DIONs, as the method relies on strong-form PDE residuals that require differentiability with respect to the coefficient field. Consequently, the coefficient decoder in PI-DIONs is designed to predict permeability values directly. To obtain binary reconstructions, a threshold of 7.5 is then applied to convert the continuous predictions into discrete phase labels.
For IGNO, the $29\times 29$ coefficient image is used as the coefficient encoder input. Since the boundary condition is fixed ($g=0$) across all samples, the boundary encoder is omitted. To enforce boundary conditions automatically, a mollifier $f(x,y)=\sin(\pi x)\sin(\pi y)$ is applied to the solution decoder output:
\begin{equation}
u_{pred}(x,y) = \mathcal{G}_{\bt_u}(\bm{\beta})(x,y)\cdot f(x,y).
\end{equation}
Below, we present the network architectures for both methods:
\paragraph{The IGNO method}
\begin{itemize}
    \item \textbf{Encoder $E_{\bt_{\bm{\beta}_1}}$.} The encoder $E_{\bt_{\bm{\beta}_1}}$ is selected as an FFCN. The input coefficient image is first flattened into a vector, then passed through two dense layers with 512 and 256 neurons, respectively. SiLU activations are applied in all hidden layers, and a Tanh activation is applied at the output.
    \item \textbf{Solution Decoder $\mathcal{G}_{\bt_u}$.} The solution decoder $\mathcal{G}_{\bt_u}$ adopts the MultiONet architecture. Both the branch and trunk networks are FFCNs with five hidden layers of 100 neurons each. The custom activation function Tanh\_Sin (Eq.~\eqref{eq:tanh_sin}) is applied to all hidden layers.
    \item \textbf{Coefficient Decoder $\mathcal{G}_{\bt_a}$.} The coefficient decoder $\mathcal{G}_{\bt_a}$ outputs the probability that a given spatial point belongs to phase 1. The recovered permeability field is obtained by thresholding the probability field at $0.5$ and mapping to $10$ (phase 1) or $5$ (phase 2). Both branch and trunk networks follow the MultiONet design, with five hidden layers of 256 neurons each. The trunk employs the custom activation
    \begin{equation}\label{eq:silu_sin}
    \text{SiLU\_Sin}(x) = \text{SiLU}(\sin(\pi x + \pi)) + x,
    \end{equation}
    while the branch employs
    \begin{equation}\label{eq:silu_id}
    \text{SiLU\_Id}(x) = \text{SiLU}(x) + x.
    \end{equation}
    A Sigmoid activation is applied at the output to ensure predictions lie in $[0,1]$.
    \item \textbf{NF model $F_{\bt_{NF}}$.} The NF model consists of three flow steps. Each step is parameterized by an FFCN with two hidden layers of 128 neurons and SiLU activations.
\end{itemize}
\paragraph{The PI-DIONs method}
\begin{itemize}
    \item \textbf{Decoder $\mathcal{G}_{\bt_u}$.} The solution decoder  of PI-DIONs adopts the same architecture as in IGNO. However, its branch network takes as input the solution-based measurements $\mathcal{M}_{sol}$ rather than the latent variable $\bm{\beta}$.
    \item \textbf{Decoder $\mathcal{G}_{\bt_a}$.} Unlike IGNO, PI-DIONs predict the coefficient field $a$ directly, because probability-based cut-offs are non-differentiable and incompatible with strong-form residual training. For fairness, the decoder $\mathcal{G}_{\bt_a}$ adopts the same architecture as in IGNO, except that the branch network input is $\mathcal{M}_{sol}$, and no activation is applied at the output layer.
\end{itemize}
\subsubsection{Latent-space Optimization Details}
Gradients are computed via PyTorch automatic differentiation. The weights $\rho_{data}$ and $\rho_{pde}$ are set to $1$ and $1$, respectively. The ADAM optimizer is used with an initial learning rate of $0.1$, halved every 50 steps. Each inverse problem is solved with 500 gradient updates.
\subsection{The EIT Problem with Operator-based Measurements}
\subsubsection{Problem Setup}
The EIT is governed by the elliptic PDE with Dirichlet boundary conditions:
\begin{equation}
\begin{aligned}
-\nabla(\gamma(x,y)\nabla u(x,y)) &= 0, \quad \text{in} \ \Omega=[0,1]^2, \\
u(x,y) &= g(x,y),\quad \text{in}\ \partial\Omega, 
\end{aligned}
\end{equation}
where $\gamma>0$ denotes the unknown conductivity field, assumed to be smooth, and $g$ is the prescribed Dirichlet voltage. The goal is to recover $\gamma$ from the DtN map $\Lambda_\gamma$, defined as:
\begin{equation}
    \Lambda_\gamma[g]: g \longrightarrow \gamma\frac{\partial u}{\partial\vec{n}}|_{\partial\Omega},
\end{equation}
which maps the input voltage $g$ into the current $\gamma\frac{\partial u}{\partial\vec{n}}=\gamma\nabla u\cdot\vec{n}$ at boundary with $\vec{n}$ being the unit outward normal vector. 
In practice, the DtN map is approximated using a finite set of pre-defined voltage–current pair, defined as $\mathcal{M}_{op}=\left\{\left(g_l(\bm{X}_m), \Lambda_\gamma[g_l](\bm{X}_m)\right)\right\}^{L}_{l=1}$, where $\bm{X}_m=(
\bm{x}_1,\cdots,\bm{x}_m
)$ denotes sensors on the boundary. In our experiments, we use $m=128$ equally spaced sensors along the four boundaries, as illustrated on the right of Fig.~\ref{fig:EIT_example}(a). To approximate the DtN map, we set $L=20$, with the $l$-th input voltage defined as $\cos(2\pi(x\cos(\theta_l) + y\sin(\theta_l)))$ with $\theta_l=2\pi l/20$. Fig.~\ref{fig:EIT_example}(b) shows examples of three input voltages ($l=1,10,20$), their corresponding boundary currents, and the PDE solutions, with the conductivity field displayed on the left of Fig.~\ref{fig:EIT_example}(a).

Since the EIT problem involves operator-based measurements, the PI-DIONs method is not applicable. Therefore, we only evaluate the proposed IGNO in this problem. For training, $N=1000$ conductivity fields are generated from trigonometric functions of the form $\gamma(x,y)=\sum^{K}_{k=1}\exp(c_k\sin(k\pi x)\sin(k\pi y))$ with $K=mod(\bar{K})$, where $\bar{K}\sim\text{U}[1,5]$ and $c_k\sim\text{U}[-1,1]$. An example of a sampled conductivity is shown on the left of Fig. \ref{fig:EIT_example}(a). For each conductivity, the $L=20$ input voltages $g_l$ are used as Dirichlet conditions in \eqref{eq:EIT}, leading to $NL=20000$ total data samples $\{(\gamma^{(n)}, g^{(n)})\}^{NL}_{n=1}$. Importantly, the training of IGNO does not require computing the corresponding currents, avoiding expensive PDE simulations.

For the inverse problem, we consider two target scenarios:
\begin{itemize}
    \item\textbf{In-distribution target:} $\gamma$ is drawn from the same distribution used for training.
    \item\textbf{Out-of-distribution target:} $\gamma$ is sampled from a shifted distribution with $\bar{K} \sim \text{U}[1,5]$ and $c_k \sim \text{U}[1,1.5]$.
\end{itemize} 
In both cases, operator-based measurements $\mathcal{M}_{op}$ are generated by solving \eqref{eq:EIT} with FEM and adding Gaussian noise. We consider three noise levels: SNR$=50$ (low), SNR$=25$ (medium), and SNR$=15$ (high).

\subsubsection{Model Setup}\label{sec:model_EIT}
To extract latent representations from input coefficients, a uniform grid $\Xi_a$ with size $32\times 32$ is used, which leads to the coefficient encoder input being $\bm{a}=a(\Xi_a)$. Since this problem involves multiple boundary conditions, a boundary encoder $E_{\bt_{\bm{\beta}_2}}$ is needed to encode boundary inputs. Given that only $L=20$ boundary conditions are used, this encoder is implemented as a one-hot encoding rather than a trainable network. To enforce boundary conditions automatically, we define a mollifier for the solution decoder output. Specifically, for solution prediction with boundary condition $g_l$:
\begin{equation}
u_{pred} = \mathcal{G}_{\bt_u}(\bm{\beta}) \cdot f + g_l ,
\end{equation}
where $f(x,y) = \sin(\pi x)\sin(\pi y)$. This ensures that the predicted solution satisfies the boundary conditions without introducing an explicit boundary loss term. 

Below, we provide details of the network architectures:
\begin{itemize}
    \item \textbf{Coefficient Encoder $E_{\bt_{\bm{\beta}_1}}$.} The coefficient encoder $E_{\bt_{\bm{\beta}_1}}$ consists of a CNN followed by an FFCN. The CNN has four hidden layers with 64 output channels per layer, kernel size $(3,3)$, and a stride of 2. The subsequent FFCN has two hidden layers of 64 neurons each. SiLU activations are applied to all hidden layers, and Tanh is used at the output.
    \item \textbf{Boundary Encoder $E_{\bt_{\bm{\beta}_2}}$.} To encode the boundary conditions $\{g_l\}_{l=1}^L$, we use a one-hot encoding: for boundary $g_l$, the latent vector $\bm{\beta}_2 = e_l \in \mathbb{R}^{20}$, where $e_l$ is the $l$-th unit vector. This enables IGNO to handle operator-based measurements with multiple boundary conditions efficiently.
    \item \textbf{Solution Decoder $\mathcal{G}_{\bt_u}$.}  The solution decoder adopts the MultiONet architecture. Both branch and trunk networks are FFCNs with five hidden layers of 100 neurons per layer. The custom Tanh\_Sin activation is applied to all hidden layers.
    \item \textbf{Coefficient Decoder $\mathcal{G}_{\bt_a}$.} The coefficient decoder $\mathcal{G}_{\bt_a}$ shares the same architecture as $\mathcal{G}_{\bt_u}$.
    \item \textbf{The NF model $F_{\bt_{NF}}$.} The NF model consists of three flow steps, each parameterized by an FFCN with two hidden layers of 128 neurons and SiLU activations.
\end{itemize}

\subsubsection{Latent-space Optimization Details}
For In-distribution targets, gradients of the objective $\mathcal{F}$ with respect to $\bm{\beta}_1$ are computed via PyTorch automatic differentiation. Weights are set as $\rho_{data}=100$ and $\rho_{pde}=1$. The ADAM optimizer is used with an initial learning rate of $0.01$, reduced by half every 25 steps, for a total of 200 updates.
For Out-of-distribution targets, weights are set to $\rho_{data}=100$ and $\rho_{pde}=0.001$. The ADAM optimizer is used with an initial learning rate of $0.01$, decreased by a factor of $2/3$ every 100 steps, with a total of 200 updates.

\section{Supplementary Tables}
\subsection{Continuous Coefficient Recovery with Solution-based Measurements}
\subsubsection{Table 1: In-distribution case} 
IGNO achieves 3 to 7 times lower reconstruction errors than the PI-DIONs method across all noise levels for in-distribution continuous coefficient recovery. Even under severe noise (SNR=15 dB), IGNO maintains RMSE below 1.2\%, while the PI-DIONs method exceeds 4\%.
\begin{table}[!htbp]\small
\centering
\caption{RMSEs obtained by different methods under different noise levels in recovery of \textbf{continuous coefficients} with solution-based measurements. (\textbf{In-distribution case})}
\begin{tabular}{c|c|c|cc} \bottomrule
                    {} & $\text{SNR}=50$ & $\text{SNR}=25$ & $\text{SNR}=15$ \\ \hline
    {IGNO} & \textbf{$0.0056$} & $0.0061$  & $0.0115$ \\
    {PI-DIONs} & \textbf{$0.0366$} & $0.0380$ & $0.0415$ \\\toprule
\end{tabular}
\label{tab:darcy_smh_in}
\end{table}

\subsubsection{Table 2: Out-of-distribution case} 
For out-of-distribution targets with frequency parameters entirely outside training range, IGNO maintains strong generalization with only modest performance degradation compared to in-distribution cases. The PI-DIONs method shows substantial error increases (4.8-6.4\%), while IGNO remains 3 to 5 times more accurate.
\begin{table}[!htbp]\small
\centering
\caption{RMSEs obtained by different methods under different noise levels in recovery of \textbf{continuous coefficients} with solution-based measurements. (\textbf{Out-of-distribution case})}
\begin{tabular}{c|c|c|cc} \bottomrule
                    {} & $\text{SNR}=50$ & $\text{SNR}=25$ & $\text{SNR}=15$ \\ \hline
    {Inv-GenNO} & \textbf{$0.0064$} & $0.0113$  & $0.0194$ \\
    {PI-DIONs} & \textbf{$0.0476$} & $0.0484$ & $0.0639$ \\\toprule
\end{tabular}
\label{tab:darcy_smh_out}
\end{table}
\subsection{Piecewise-Constant Coefficient Recovery with Solution-based Measuremen}
\subsubsection{Table 3: In-distribution case} 
The cross-correlation indicator $I_{corr}$ measures morphological similarity, with values near 1 indicating strong agreement between reconstructed and true phase topologies. IGNO achieves $I_{corr} > 0.95$ across all noise levels, indicating accurate phase topology recovery. The PI-DIONs method fails with $I_{corr}\sim 0.72$, indicating essentially random reconstructions due to the inability to handle undefined gradients at discontinuities.
\begin{table}[!htbp]\small
\centering
\caption{Cross-correlations $I_{corr}$ obtained by different methods under different noise levels in recovery of \textbf{piecewise-constant coefficients} with solution-based measurements. (\textbf{In-distribution case})}
\begin{tabular}{c|c|c|cc} \bottomrule
                    {} & $\text{SNR}=50$ & $\text{SNR}=25$ & $\text{SNR}=15$ \\ \hline
    {IGNO} & \textbf{$0.969$} & $0.960$  & $0.951$ \\
    {PI-DIONs} & \textbf{$0.718$} & $0.715$ & $0.726$ \\\toprule
\end{tabular}
\label{tab:darcy_pwc_in}
\end{table}
%
\subsubsection{Table 4: Out-of-distribution case} 
For out-of-distribution piecewise-constant targets with different Gaussian process correlation structures producing more complex phase geometries, IGNO maintains $I_{corr}> 0.91$ even under high noise. The PI-DIONs method continues to fail ($I_{corr}\sim 0.78$), demonstrating an inability to generalize for discontinuous inverse problems.
\begin{table}[!htbp]\small
\centering
\caption{Cross-correlations $I_{corr}$ obtained by different methods under different noise levels in recovery of \textbf{piecewise-constant coefficients} with solution-based measurements. (\textbf{Out-of-distribution case})}
\begin{tabular}{c|c|c|cc} \bottomrule
                    {} & $\text{SNR}=50$ & $\text{SNR}=25$ & $\text{SNR}=15$ \\ \hline
    {IGNO} & \textbf{$0.960$} & $0.949$  & $0.914$ \\
    {PI-DIONs} & \textbf{$0.778$} & $0.779$ & $0.777$ \\\toprule
\end{tabular}
\label{tab:darcy_pwc_out}
\end{table}

\subsection{The EIT Problem with Operator-based Measurements}
\subsection{Table 5: the EIT problem}
For the operator-based EIT problem, IGNO demonstrates robust performance across both distribution types. In-distribution reconstruction achieves sub-0.5\% error even under low noise, degrading gracefully to 2.8\% at SNR=15 dB. Out-of-distribution targets show only modest error increases (2.2-2.9\%), indicating strong extrapolation capabilities. The PI-DIONs method cannot handle operator-based measurements and does not apply to this problem.
\begin{table}[!htbp]\small
\centering
\caption{RMSEs obtained by the proposed IGNO in solving the \textbf{EIT} problem in both In-distribution and Out-of-distribution cases under different noise levels.}
\begin{tabular}{c|c|c|cc} \bottomrule
                    {} & $\text{SNR}=50$ & $\text{SNR}=25$ & $\text{SNR}=15$ \\ \hline
    {In-distribution} & \textbf{$0.0044$} & $0.0128$  & $0.0277$ \\
    {Out-of-distribution} & \textbf{$0.0219$} & $0.0246$ & $0.0293$ \\\toprule
\end{tabular}
\label{tab:EIT_smh}
\end{table}

\section{Supplementary Figures}
\subsection{Figure 1: The MultiONet architecture}
\begin{figure}[!htbp]
\centering
\includegraphics[width=1.\textwidth]{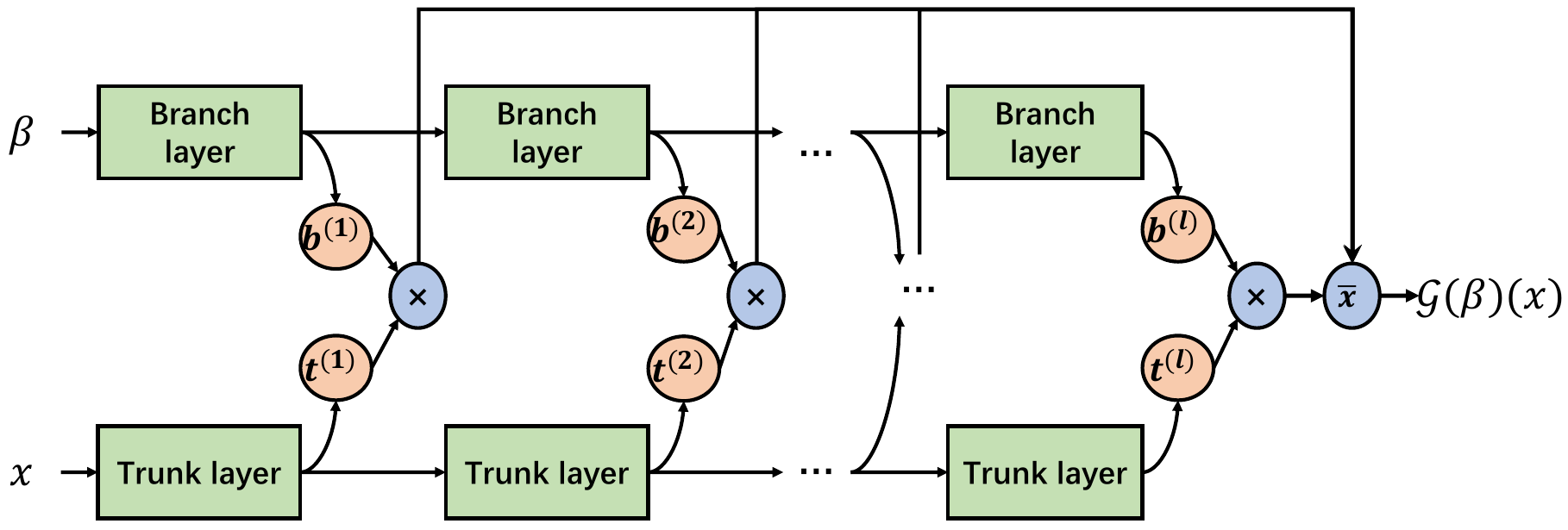}
\caption{The MultiONet architecture.}
\label{fig:MultiONet}
\end{figure}

The MultiONet architecture in Fig.~\ref{fig:MultiONet} shows branch and trunk networks with multi-layer feature aggregation.

\subsection{Figure 2: Illustration of operator-based measurements in the EIT problem}
\begin{figure}[!htbp]
\centering
\includegraphics[width=0.65\textwidth]{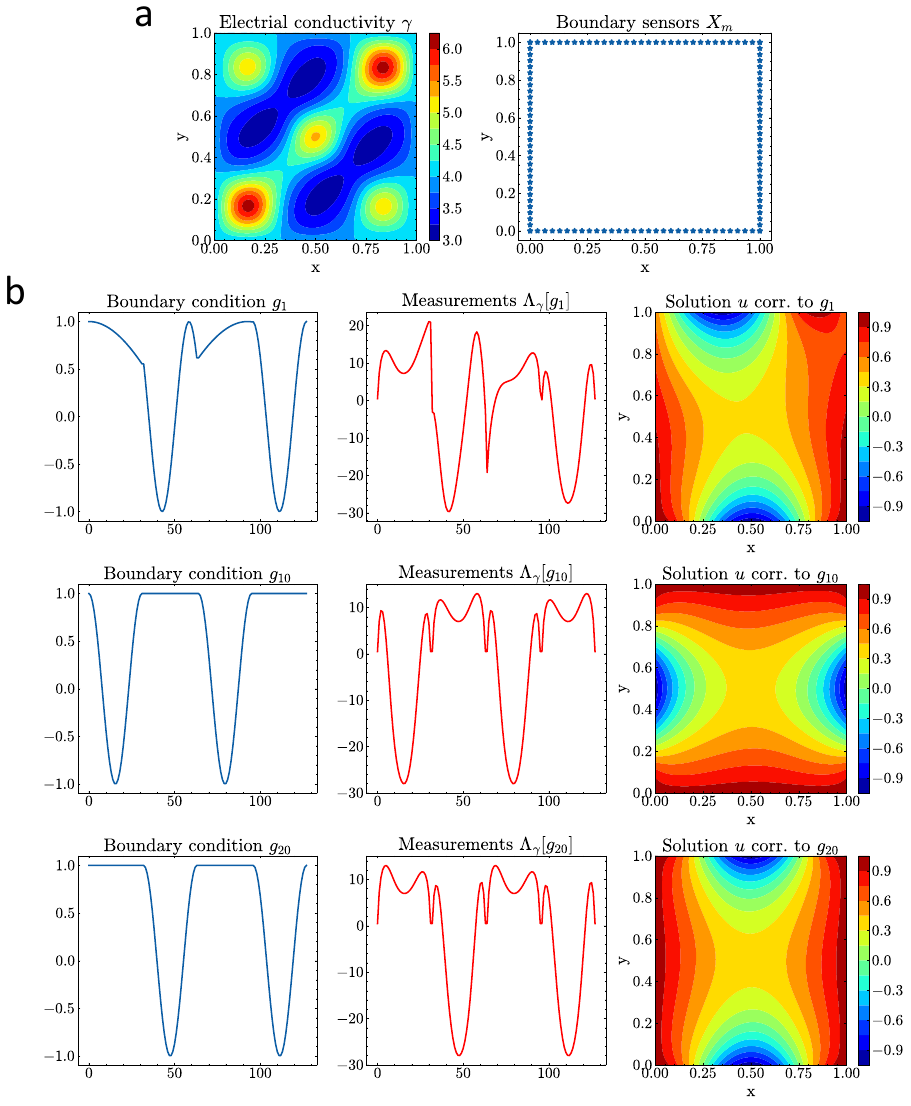}
\caption{Illustration of operator-based measurements in the EIT problem: (a) An example of conductivity $\gamma$ (left) and the boundary sensors $\bm{X}_m$ (right); (b) Examples of boundary conditions $g_l$ (left),  corresponding current measurements $\Lambda_{\gamma}[g_l]$ (middle), and PDE solutions $u$ (right), when $l=1,10,20$.}
\label{fig:EIT_example}
\end{figure}

In the EIT problem, the unknown conductivity field $\gamma$ is inferred from the DtN operator $\Lambda_{\gamma}$, which maps imposed boundary voltages $g$ to the corresponding boundary currents $\gamma\frac{\partial u}{\partial \vec{n}}|_{\partial\Omega}$. This operator is approximated using $L = 20$ distinct boundary voltage patterns and measurements collected from $m = 128$ equally spaced boundary sensors. Figure~\ref{fig:EIT_example}(a) illustrates an example conductivity field $\gamma$ (left) and the corresponding boundary sensor configuration (right). Figure~\ref{fig:EIT_example}(b) shows three representative boundary conditions $g_l$, the resulting current responses $\Lambda_{\gamma}[g_l]$, and the corresponding PDE solutions $u$ for $l = 1, 10, 20$. 

\end{document}